\documentclass{article}

\usepackage{microtype}
\usepackage{graphicx}
\usepackage{subfigure}
\usepackage{booktabs} 
\usepackage{siunitx}
\usepackage{dirtytalk}
\usepackage{hyperref}

\usepackage[accepted]{icml2024}

\usepackage{amsmath}
\usepackage{amssymb}
\usepackage{mathtools}
\usepackage{amsthm}
\usepackage{tcolorbox}
\usepackage[capitalize,noabbrev]{cleveref}
\definecolor{mBlue}{HTML}{173b4f} 
\theoremstyle{plain}

\theoremstyle{definition}

\theoremstyle{remark}

\usepackage[textsize=tiny]{todonotes}

\icmltitlerunning{Injecting Ecological Priors from Large Language Models into Neural Networks}

\begin{document}

\twocolumn[
\icmltitle{Human-like Category Learning by Injecting Ecological Priors from Large Language Models into Neural Networks}



\icmlsetsymbol{equal}{*}

\begin{icmlauthorlist}

\icmlauthor{Akshay K. Jagadish}{mpi,cch}
\icmlauthor{Julian Coda-Forno}{mpi,cch}
\icmlauthor{Mirko Thalmann}{mpi,cch}
\icmlauthor{Eric Schulz}{equal,mpi,cch}
\icmlauthor{Marcel Binz}{equal,mpi,cch}
\end{icmlauthorlist}

\icmlaffiliation{mpi}{Computational Principles of Intelligence Lab, Max Planck Institute for Biological Cybernetics, Tübingen, Germany}
\icmlaffiliation{cch}{Institute for Human-Centered AI, Helmholtz Computational Health Center, Munich, Germany}

\icmlcorrespondingauthor{Akshay K. Jagadish}{akshaykjagadish@gmail.com}

\icmlkeywords{Machine Learning, ICML}

\vskip 0.3in
]



\printAffiliationsAndNotice{\icmlEqualContribution} 

\begin{abstract} 

Ecological rationality refers to the notion that humans are rational agents adapted to their environment. However, testing this theory remains challenging due to two reasons: the difficulty in defining what tasks are ecologically valid and building rational models for these tasks. In this work, we demonstrate that large language models can generate cognitive tasks, specifically category learning tasks, that match the statistics of real-world tasks, thereby addressing the first challenge. We tackle the second challenge by deriving rational agents adapted to these tasks using the framework of meta-learning, leading to a class of models called \emph{ecologically rational meta-learned inference} (ERMI). ERMI quantitatively explains human data better than seven other cognitive models in two different experiments. It additionally matches human behavior on a qualitative level: (1) it finds the same tasks difficult that humans find difficult, (2) it becomes more reliant on an exemplar-based strategy for assigning categories with learning, and (3) it generalizes to unseen stimuli in a human-like way. Furthermore, we show that ERMI's ecologically valid priors allow it to achieve state-of-the-art performance on the OpenML-CC18 classification benchmark.

\end{abstract}

\section{Introduction} \label{introduction}

   Ecological rationality refers to the idea that humans are \emph{rational} agents adapted to the \emph{ecological} environments they interact with.
   Nearly seventy years ago, \citet{brunswik1955representative} emphasized that we have to move beyond laboratory settings and understand cognition in the light of naturalistic environments. Later on, \citet{simon1990invariants} famously argued that human decision-making is like the two blades of a scissor, with one blade representing the cognitive processes of the mind and the other the structure of the environment in which the mind operates. \citet{Todd2012-ta} furthered this notion by introducing the term ecological rationality, suggesting that minds are adapted to their environments through the use of simple, context-specific strategies.

   However, it has remained challenging to build computational models that describe how people implement strategies adapted to their environment for two reasons. First, defining ecologically valid tasks is still an open problem \cite{barker1968ecological, neisser1987cognition, hammond1998ecological} and second, even if we have access to such tasks, it is challenging to build models that solve them rationally. 

   In the present paper, we address both of these challenges. We show that large language models (LLMs) -- having been trained on large amounts of human-generated text  -- can serve as a useful tool for generating ecologically valid tasks, thereby addressing the first challenge. To address the second challenge, we then derive rational learning algorithms for these tasks using the framework of meta-learning \cite{Pratt1998-gq, hochreiter2001learning, binz2023meta}, leading to a class of models that we call \emph{ecologically rational meta-learned inference} (ERMI).

    We illustrate our approach using the domain of category learning \cite{ashby_human_2005} --- one of the best-studied areas of cognitive science. We begin by verifying that LLMs can generate category learning tasks whose statistics match real-world classification data sets \cite{oml-benchmarking-suites}. Following this, we show that ERMI quantitatively explains human data from two different category learning experiments better than seven other cognitive models. Furthermore, ERMI aligns with human behavior qualitatively: (1) it finds the same tasks difficult that humans find difficult, (2) it shows the same transition of categorization strategies as humans, and (3) it generalizes to unseen stimuli in a human-like way.
    Taken together, these results suggest that we can explain human category learning to a large extent using the principle of ecological rationality.

    Furthermore, we hypothesized that the ecologically valid priors encoded in ERMI allow it to perform well on classification tasks from the machine learning literature. To test this hypothesis, we evaluate ERMI on the curated classification benchmark OpenML-CC18 \cite{oml-benchmarking-suites} and find that it achieves state-of-the-art performance. 

\section{Related work}
    
    
    \textbf{LLMs for data generation:} 
    Recently, the wider concept of using LLM-generated data to train another model has become more popular \cite{gunasekar2023textbooks, schick2021generating,  wang2023improving, bai2022constitutional, mitra2023orca, alpaca}. For example, \citet{gunasekar2023textbooks} prompted GPT-3.5 to generate synthetic textbook-quality data which they used to train a smaller transformer-based model. To justify this approach in the context of ecological rationality, one has to first establish that LLMs can produce ecologically valid tasks. \citet{Borisov2022-sr} have done so recently by showing that LLMs are realistic tabular data generators, while \citet{Griffiths2023-pt} demonstrated that LLM-generated data matches the priors of human subjects in several settings. \cite{coda2023meta} have shown that LLMs can even adapt their priors by meta-learning fully in-context. 

    \textbf{Meta-learned models of cognition:} 
    Using models that achieve optimal task performance to study behavior is central to the rational analysis of cognition \cite{anderson1991human}. Traditionally, these models have taken the form of Bayesian models \cite{l2008bayesian}. However, the Bayesian framework does not permit the construction of rational models for a given data set of tasks. The framework of meta-learning offers a way to overcome this problem \cite{binz2023meta}. Unlike Bayesian models, meta-learned models of cognition can learn adaptive priors by repeatedly interacting with a distribution of tasks. Furthermore, these models have been shown to converge onto the optimal learning algorithm for the environments they are trained on \cite{ortega2019meta} and can be used in cases where the hand-crafting of assumptions is impractical or even infeasible. 
    
    Recently, it has been shown that meta-learned models capture human behavior across a wide range of domains, including decision-making \cite{binz2022heuristics}, reinforcement learning \cite{kumar2022using, binz2022exploration, jensen2023recurrent, schubert2023rational}, and compositional reasoning \cite{jagadish2023zero, lake2023human}. However, all these previous applications have relied on environments hand-engineered by researchers instead of ecologically valid ones. 
    

    \textbf{Human category learning:} How people learn to categorize objects has received significant attention in the cognitive sciences. For example, researchers have investigated how people learn to make fine-grained perceptual categorizations \cite{ashby_varieties_1986}, what strategies people use when learning to categorize objects by comparing formal models of category learning \cite{smith1998prototypes, nosofsky_exemplar_2002, maddox_comparing_1993}, or whether there are different cognitive systems of category learning \cite{ashby_human_2005, newell_chapter_2011}. In the present work, we make use of this rich literature by relying on its experimental paradigms and data. In particular, we used the paradigms developed by \citet{Shepard1961-yu}, \citet{smith1998prototypes}, and \citet{Johansen2002-xe}. We furthermore compare our model to a wide range of previously established category learning models \cite{nosofsky1986attention, Anderson1991-ii, homa1984role, Nosofsky1994-gu}.

\section{Methods}

    In this section, we describe how we prompted LLMs to generate ecologically valid category learning tasks and how we then used meta-learning to learn models that are optimally adapted to these tasks.

    \subsection{Prompting LLMs to generate ecologically valid category learning tasks} \label{sec:sampling}

    A category learning task entails categorizing a stimulus $x \in \mathbb{R}^n$ into categories $y$ based on its feature values.  Multiple stimuli are presented sequentially and participants are tasked to predict their category after each presentation. Upon making their choice, they receive feedback on the true category of the stimulus and are presented with the next one.

    To generate thousands of such category learning tasks from an LLM, we relied on a two-stage process. In the first stage, we queried the LLM to synthesize feature names and corresponding category labels. In the second stage, the model was prompted to produce data points for the feature names and category labels generated in the first stage. 

    More specifically, we used the following prompt to synthesize feature names and category labels: 
    \begin{tcolorbox}[sharp corners, colback=mBlue!5!white,colframe=mBlue!75!black, width=0.48\textwidth, left=4pt,right=4pt, top=4pt, bottom=4pt, title=\textbf{Synthesize feature names and category labels} \label{prompt:task_label}]
    I am a psychologist who wants to run a category learning experiment. In a category learning experiment, there are many different three-dimensional stimuli, each of which belongs to one of two possible real-world categories.\\ 

    Please generate names for three stimulus feature dimensions and two corresponding categories for 250 different category learning experiments:
    \end{tcolorbox}

     The LLM then produced a series of category learning tasks in a sequence until the specified number of tasks was generated. To illustrate one example, based on this prompt, the model constructed a category learning task with \textsc{[sodium, fat, protein]} content as feature names and \textsc{[healthy, unhealthy]} as category labels. 

    In the second stage, we prompted the LLM to generate data points for a given category learning task: 

    \begin{tcolorbox}[sharp corners, colback=mBlue!5!white,colframe=mBlue!75!black, width=0.48\textwidth, left=4pt,right=4pt, top=4pt, bottom=4pt, title=\textbf{Generate category learning tasks} \label{prompt:task_values}]
    I am a psychologist who wants to run a category learning experiment. For a category learning experiment, I need a list of stimuli and their category labels. Each stimulus is characterized by three distinct features: {\color{mBlue}\textbf{sodium}}, {\color{mBlue}\textbf{fat}}, and {\color{mBlue}\textbf{protein}}. These features can take only numerical values. The category label can take the values {\color{mBlue}\textbf{healthy}} or {\color{mBlue}\textbf{unhealthy}} and should be predictable from the feature values of the stimulus.\\

    Please generate a list of 100 stimuli with their feature values and their corresponding category labels using the following template for each row:\\
    
    -- feature value 1, feature value 2, feature value 3, \\ \phantom{-} category label 
    \end{tcolorbox}

    Each generated data point contains feature values and their corresponding category label, e.g., \textsc{[250, 15, 20, healthy]} for our previously mentioned example. In total, we generated three data sets containing around \num{10000} different category learning tasks for three, four, and six feature dimensions. Each task consisted of $100$, $300$, and $616$ data points, respectively. We provide further details about the generated category learning tasks (including the counts of the top 50 feature names and category labels) in Appendix \ref{app:generate_tasks}.

    For our data generation procedure, we used \textsc{Claude-v2} \cite{Anthropic2023-hr} as it can process up to \num{100000} tokens, is instruction-tuned, and performed well out of the box in our preliminary experiments. The temperature parameter was set to one to induce diversity and all other parameters were set to their default values. We provide details about other LLMs we considered and additional design choices in Appendix \ref{app:prompting}.

\begin{figure*}[!h]
    \centering
    \includegraphics[width=\textwidth]{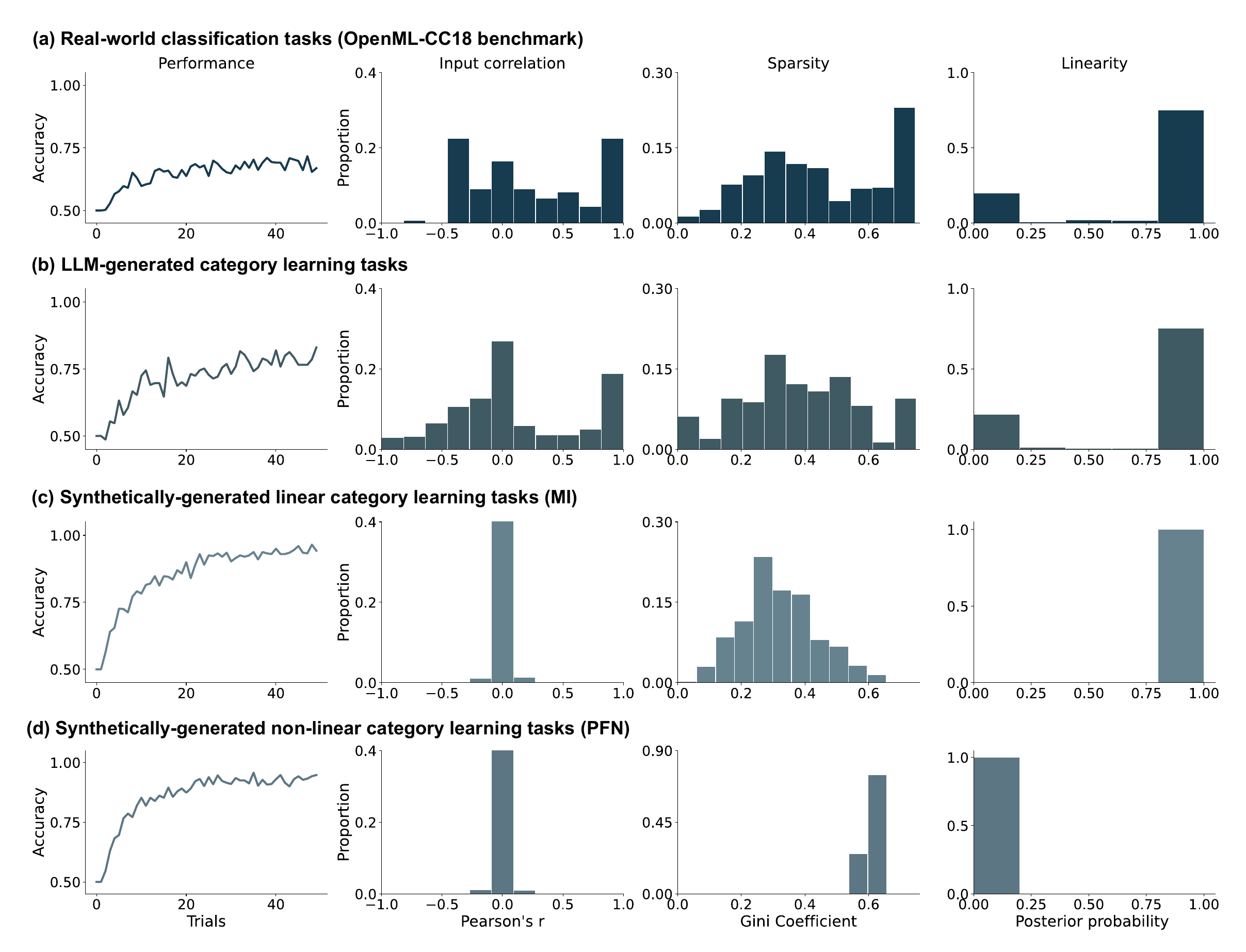}
    \begin{center}
    \caption{\textbf{LLM generates ecologically valid category learning tasks:} Mean task performance of the logistic regression model measured over trials (first column). Histogram of Pearson's correlation coefficients computed between pairs of features (second column). Histogram of Gini coefficients computed over the logistic regression weights (third column). Linearity of the category learning task (fourth column) computed for (a) 28 different real-world binary classification tasks from the OpenML-CC18 benchmarking suite (b) ecologically valid category learning tasks generated from \textsc{Claude-v2} and (c) synthetic category learning tasks derived using the Bayesian logistic regression prior that were used to train the meta-learned inference (MI) model (d) synthetic category learning tasks with nonlinear decision boundary derived using the Bayesian neural network prior that were used to train prior-fitted networks (PFN) model.}
    \label{fig:stats}
    \end{center}
\end{figure*}

        
        
    
    \subsection{Ecologically rational meta-learned inference (ERMI)}

    We parsed the generated tasks as described in Appendix \ref{app:parsing} and stored them in a numerical format. Then, we constructed rational learning algorithms for the numerical data by training memory-based meta-learning systems based on a two-stage process \cite{hochreiter2001learning, santoro2016meta, wang2016learning}. In an inner-loop stage, a neural network predicts the category $y_t$ for an input stimulus $x_t$ conditioned on preceding stimulus-category pairs $x_{1:t-1}, y_{1:t-1}$. In an outer-loop stage, the network's parameters $\theta$ are updated using the following objective:
    \begin{equation}
        \underset{\theta}{\arg \max} ~\mathbb{E}_{ p(x_{1: T}, y_{1: T})}\left[\sum_{t=1}^{T}\log p_{\theta} \left(y_{t} \mid x_{1: t}, y_{1:t-1} \right)\right]
        \label{eq:loss}
    \end{equation}
    where $p_{\theta}$ defines the output probabilities produced by the network. 
    
    During evaluation -- i.e., once training is completed -- the neural network implements a free-standing learning algorithm that can predict the category label of a new stimulus based on preceding stimulus-category pairs, despite its parameters being frozen. The resulting network approximates the Bayes-optimal learning algorithm for the data set of the category learning tasks $p(x_{1: T}, y_{1: T})$ encountered during training \cite{ortega2019meta}. 
    
    We refer to the class of models derived by training on ecologically valid (i.e., LLM-generated) category learning task as \emph{ecologically rational meta-learned inference} (ERMI). If trained on synthetically-generated category learning tasks sampled from a Bayesian logistic regression prior, we refer to the models as \emph{meta-learned inference} (MI; \citealp{binz2022heuristics}). We chose MI as a baseline for two reasons. First, linear models have a long history as models of human learning \cite{karelaia2008determinants, lucas2015rational} and also category learning more specifically \cite{speekenbrink2008learning, speekenbrink2010models}. Furthermore, the model has been previously used in a setting known as multiple cue probability learning \cite{binz2022heuristics}. While not exactly the same, this setting shares many similarities to category learning. Finally, using the terminology of \citet{muller2022transformers}, we refer to the models as \emph{prior-data fitted networks} (PFN) when tasks are sampled from a Bayesian neural network prior. For details on how these tasks are generated, see Appendix \ref{app:synthetic_data}.
    
     The backbone for all our meta-learning models consisted of a transformer-based decoder architecture \cite{vaswani2017attention} with a causal attention mask. The network had six layers, a model dimension of 64, 256 hidden units in the feed-forward network, and eight attention heads. Positional encoding of input data points was done using sine and cosine functions of different frequencies \cite{vaswani2017attention}. Note that during evaluation, transformer weights are frozen and learning is purely driven by self-attention applied to causally masked inputs. 

    In each training episode, a batch of tasks is sampled from $p(x_{1: T}, y_{1: T})$ and the model predicts the category for the given stimulus conditioned on all preceding stimulus-category pairs. Finally, the objective mentioned in Equation \ref{eq:loss} is computed, and model parameters are updated using the \textsc{Adam} optimizer \cite{kingma2014adam} with a learning rate of \(10^{-4}\). This process is repeated for \num{500000} episodes. We provide full details about the model training procedure in Appendix \ref{app:model_training}.

\section{LLM-generated category learning tasks are ecologically valid} \label{sec:llm_stats}

    ERMI can only be interpreted as an ecologically rational model if the statistics of LLM-generated tasks on which it was trained match the statistics of real-world classification problems. We verified that this was the case by comparing the data distributional properties of the two \cite{chan2022data}. For this analysis, we relied on the OpenML-CC18 benchmarking suite, a curated collection of real-world classification tasks \cite{oml-benchmarking-suites}. Note that although the OpenML-CC18 benchmark is large enough to enable this form of statistical analysis, it is too small for direct meta-learning.
        
    We downsampled all tasks in the OpenML-CC18 benchmark to four feature dimensions and included only binary classification tasks without any missing features in our analysis -- amounting to 28 tasks. In addition, we also contrasted LLM-generated tasks to a collection of synthetically-generated category learning tasks with a linear decision boundary (corresponding to those used to train MI; refer to Appendix \ref{app:synthetic_data}). We analyzed these collections of tasks in terms of their learning curves, input correlations, sparsity, and linearity -- details for which can be found in Appendix \ref{app:real_stats}.
    
    \begin{figure*}[!t]
        \centering
        \includegraphics[width=\textwidth]{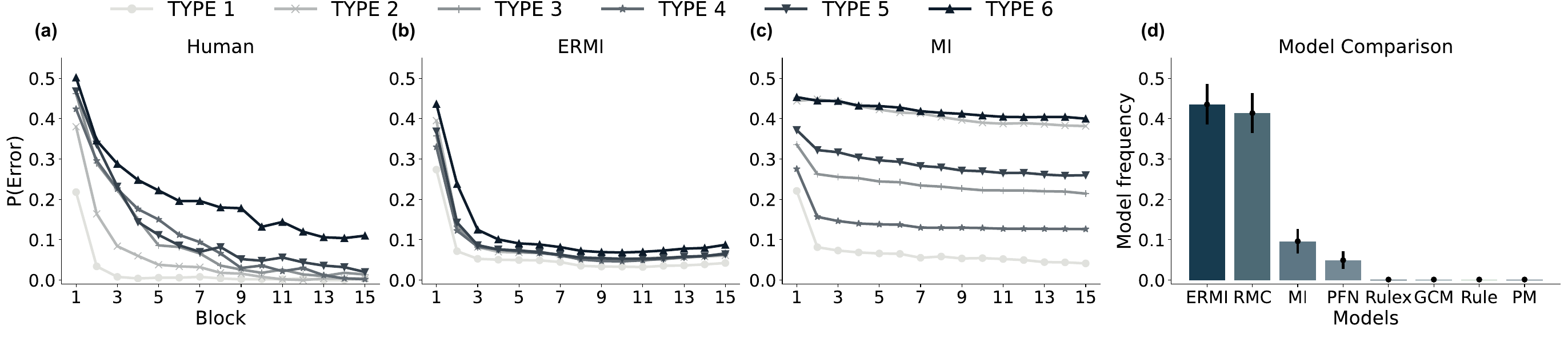}
        \caption{\textbf{ERMI shows human-like learning difficulties}: \textbf{(a-c)} Average error probabilities for each task \textsc{type} in each block of 16 trials for (a) humans, (b) ERMI, and (c) MI. \textbf{(d)} The posterior model frequency of participants' choices in the \citet{Badham2017-hc} study for eight computational models. Human data in (a) was reproduced from Table 1 in \citet{Nosofsky1994-hw}. ERMI and MI were simulated on \textsc{type 1-6} tasks for 50 runs with the inverse temperature that resulted in the lowest mean-squared error compared to humans, which was $\beta=0.4$ for ERMI, and $\beta=0.9$ for MI.}
        \label{fig:experiment1}
    \end{figure*}
    
    \textbf{Learning curves:} Real data is noisy and not perfectly predictable. To investigate whether this is also true for our LLM-generated category learning tasks, we plotted the learning curves of a logistic regression model as a function of the number of training points. We found that the model reaches a ceiling accuracy of around 75\% for both LLM-generated and real-world classification tasks (Figure \ref{fig:stats}; first column). In contrast, the ceiling performance for synthetically-generated tasks was much higher, reaching almost 100\%.
    
    \textbf{Input correlations:} Information contained in different feature dimensions are often correlated with each other. In the context of human cognition, it has been argued that this data distributional property of real-world data supports the reliance of people on heuristic decision-making strategies \cite{gigerenzer2011heuristic}. While input dimensions in the synthetically-generated data were not correlated at all, both LLM-generated ($0.11 \pm 0.02;\, t(1639) = 4.55,\, p < 0.001$) and real-world tasks $(0.21 \pm 0.01;\, t(2252) = 14.64,\, p < 0.001)$ showed a significant percentage of correlated features (Figure \ref{fig:stats}; second column). The corresponding histograms had similar shapes, both containing a peak at perfectly correlated features.

    \textbf{Sparsity:} Another data distributional property that allows people to ignore information is sparsity --- for many tasks only a few dimensions are relevant. We fitted a linear model on each task to evaluate whether we could find evidence for this in the LLM-generated data. We used the Gini coefficient -- a measure borrowed from the economics literature -- of the resulting regression coefficients to quantify sparsity \cite{binz2022heuristics}. High Gini coefficients correspond to maximal sparsity, meaning only a single feature is relevant. Both LLM-generated ($0.38 \pm 0.01;\, t(545) = 3.81,\, p < 0.001$) and real-world tasks ($ 0.45 \pm 0.01;\, t(762) = 10.83,\, p < 0.001$) exhibited significantly higher sparsity than synthetically-generated tasks ($0.32 \pm 0.01$, see Figure \ref{fig:stats}; third column).
     
    \textbf{Linearity:} People have strong priors towards linear relationships but can also learn non-linear ones given enough examples \cite{lucas2015rational, brehmer1974hypotheses}. To measure whether this bias is also present in the distributional properties of the data, we conducted a model comparison between a linear model (a simple logistic regression model) and a non-linear one (logistic regression with higher-order polynomial features). For each task, we computed the posterior probability that the linear model offers a better explanation as our measure of linearity (details can be found in Appendix \ref{app:real_stats}). Most LLM-generated and real-world tasks were found to be linear but there was also a significant number of exceptions (Figure \ref{fig:stats}; fourth column). The synthetically-generated tasks, on the other hand, were fully linear by design.

    Taken together, these analyses indicate that category learning tasks generated by LLMs share many features with real-world classification tasks. As they can also be produced in large quantities, these tasks can serve as a substitute for real-world classification tasks when meta-learning ecologically rational learning algorithms.

\section{ERMI shows human-like learning difficulties} \label{sec:experiment1}

      In the following sections, we investigate how well ERMI captures human category learning. We began by looking at one of the most canonical studies in category learning originally conducted by \citet{Shepard1961-yu}. The study required participants to learn how to categorize a stimulus that varied in shape (triangle or square), size (small or big), and color (black or white). In total, there were eight stimuli, and participants had to categorize them into one of two categories over several blocks of 16 trials. The authors assigned a stimulus to a category based on six different rules (labeled \textsc{type 1} to \textsc{type 6}) that increased in difficulty. For more information, we refer to Appendix \ref{app:experiment1}. 
     
     \citet{Shepard1961-yu} showed that people find tasks belonging to \textsc{type 1} the easiest and \textsc{type 6} the hardest, with the average error for \textsc{type 1} tasks going to zero after four blocks and \textsc{type 6} tasks remaining at around $10.6\%$ even after 15 blocks (see Figure \ref{fig:experiment1}). The average error for \textsc{type 2} tasks ($3.2\%$) was found to be lower than for \textsc{type 3} tasks ($6.1\%$), \textsc{type 4} tasks ($6.5\%$), and \textsc{type 5} tasks ($7.5\%$). 

     We simulated ERMI and MI on the \citet{Shepard1961-yu} study. Figure \ref{fig:experiment1} (a) to (c) shows their learning curves alongside those of humans for the six difficulty levels. 
     It can be seen that the learning curves of ERMI are difficulty-dependent and are in terms of mean-squared error (MSE) more similar to humans ($\mathrm{MSE} = 0.03$) than MI ($\mathrm{MSE} = 0.26$). Notably, ERMI, like humans, finds the \textsc{type 1} task easier than the \textsc{type 6} task and shows a similar clustering of learning curves for \textsc{type 2} to \textsc{type 5} tasks (see Table \ref{table:shepardtask} for details). However, we also find that ERMI learns much faster than people: its performance plateaus after four blocks while humans continue learning until the end of the experiment for most types.
     Even though MI performs tasks of different difficulty levels to varying degrees of success, its learning curves do not match those of humans. For example, unlike humans, MI finds \textsc{type 2} tasks as difficult as \textsc{type 6} tasks and \textsc{type 4} tasks easier than \textsc{type 3} tasks.  
        
    Furthermore, we investigated how well ERMI explains human trial-by-trial choices on a quantitative level. For this, we considered human data from \citet{Badham2017-hc} who conducted a replication of Shepard's original study that only included \textsc{type 1} to \textsc{type 4} tasks. We performed a Bayesian model comparison between eight computational models: the three meta-learned models introduced earlier (ERMI, MI, and PFN), and five other cognitive models.
    The five established category learning models from the cognitive science literature included the rational model of categorization (RMC; \citealp{Anderson1991-ii}), the generalized context model (GCM; \citealp{nosofsky1986attention}), a prototype model (PM; \citealp{homa1984role}), a rule-based model (Rule; \citealp{ashby_varieties_1986}), and a rule-plus-exception model (Rulex; \citealp{Nosofsky1994-gu}). 
    We provide more details about the five cognitive models, their fitting procedure, and the model comparison in Appendix \ref{app:cogmodels}, \ref{app:fitting}, and \ref{app:comparison}. 

    We measured the goodness-of-fit to human choices based on two metrics: posterior model frequency and exceedance probability \citep{rigoux2014bayesian}. The posterior model frequency measures how often a model offers the best explanation in the population, while the exceedance probability measures how likely it is that a given model is the most frequent explanation (the latter is reported in the Appendix \ref{app:experiment1}). 
    Figure \ref{fig:experiment1} (d) shows that ERMI explains human choices the best more frequently $(0.43 \pm 0.05)$ compared to the other models, with the RMC coming in a close second $(0.41 \pm 0.05)$. 
    MI $(0.10\pm 0.03)$ and PFN $(0.05 \pm 0.02)$ fit human data the best less than $10 \%$ of the time. The classical cognitive models like the exemplar-, prototype- and rule-based models failed at explaining human choices better than other competing models ($\leq 1\%$ of times).

     Additionally, we simulated \textsc{Claude-v2} directly on the \citet{Badham2017-hc} study (see Appendix \ref{app:experiment1} for details). We first observed that \textsc{Claude-v2} can perform the task optimally quite rapidly, with a mean error reaching zero for most difficulty levels already after three blocks. However, we also observed that the ordering of difficulty levels in \textsc{Claude-v2} does not match humans, with \textsc{Claude-v2} finding type 2 tasks more difficult than type 3, 4, and 5 tasks as shown in Figure \ref{fig:llmsimulationsbadham2017}. Furthermore, we found that the LLM’s (MSE=$0.18$) match to human error rates was worse than that of ERMI (MSE=$0.05$), see Table \ref{table:shepardtaskwithllm} for details. We also fitted the \textsc{Claude-v2} predictions to human choices from the \citet{Badham2017-hc} study and compared its performance to ERMI using the Bayesian information criterion (BIC; lower is better). We found that ERMI (38722.13) still offers a better fit to humans than \textsc{Claude-v2} (39816.90). Taken together, these results suggest that an LLM trained on the entire internet cannot explain human category learning as well as an ecologically rational meta-learning inference model.

\section{ERMI becomes more exemplar-based with learning} \label{sec:experiment2}
     \begin{figure*}[!t]
     \centering
     \includegraphics[width=\textwidth]{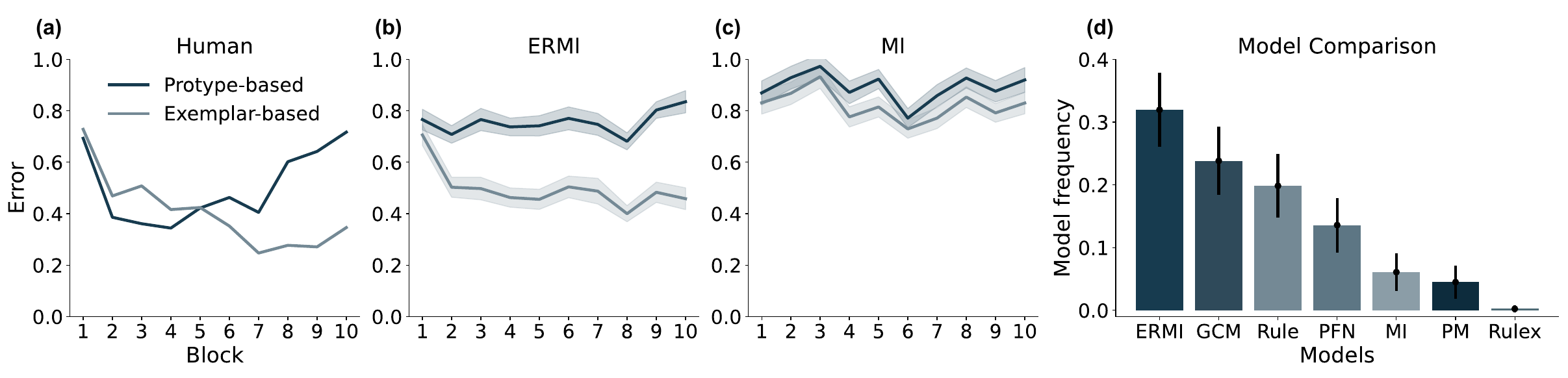}
      \caption{\textbf{ERMI becomes more exemplar-based with learning}: \textbf{(a-c)} The average error of exemplar- and prototype-based models fitted to (a) human choices, (b) simulated choices from ERMI, and (c) simulated choices from MI for each block of 56 trials. \textbf{(d)} The posterior model frequency of participants’ choices in the \citet{devraj2021dynamics} study for seven computational models. Human data in (a) was reproduced from \citet{smith1998prototypes}. ERMI and MI were simulated using inverse temperature values fitted to participants' choices in \citet{devraj2021dynamics}. The mean of the fitted inverse temperature and its standard error were $0.09 \pm 0.01$ for ERMI and $0.17 \pm 0.02$ for MI, respectively. The shaded region shows the standard error of the mean.} 
     \label{fig:experiment2}
     \end{figure*}
    What strategy people use to categorize objects and how the application of strategies changes over time are heavily debated questions in psychology. \citet{smith1998prototypes} attempted to understand whether people use prototype- or exemplar-based strategies during category learning. More specifically, they asked: do people learn a prototype for each category and assign categories based on the similarity of a stimulus to the learned prototypes, or do they instead remember previously seen examples for each category and assign categories based on the similarity of a stimulus to the stored exemplars?

    To investigate this, \citet{smith1998prototypes} designed a category learning task that contained 14 six-dimensional stimuli, each of which was assigned to a category based on a non-linear decision rule. Participants in their experiment were then tasked to assign one of the two categories to repeatedly presented stimuli. Following this, they fit predictions from prototype- and exemplar-based models to proportions of human choices, aggregated over trials within a block, by minimizing the MSE between them. They found that people were better explained by the prototype-based model in the early blocks but in the later blocks, their choices aligned more closely with the exemplar-based model as shown in Figure \ref{fig:experiment2} (a). We provide more details about this analysis in Appendix \ref{app:experiment2}.
    
    \begin{figure*}[!t]
    \centering
    \includegraphics[width=\textwidth]{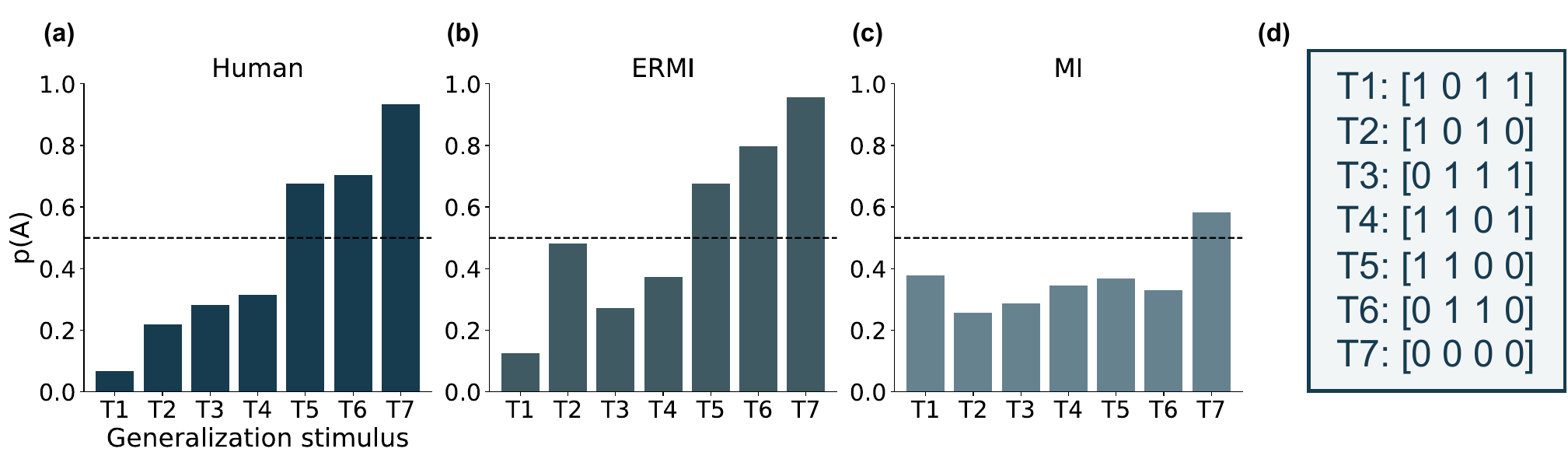}
    \caption{\textbf{ERMI displays human-like generalization:} \textbf{(a-c)} Average categorization probabilities of transfer stimuli T1-T7 for (a) humans (b) ERMI (c) MI. \textbf{(d)} The encoding scheme used for the seven transfer stimuli. Human data in (a) was reproduced from \citet{Johansen2002-xe}. ERMI and MI were simulated on the same experiment for 77 runs, with inverse temperature settings that resulted in the lowest mean-squared error compared to humans, which was $\beta=$ 0.9 for ERMI, and $\beta=$ 0.1 for MI.} \label{fig:experiment3}
    \end{figure*}
    
    We simulated choices from ERMI and MI on the task of \citet{smith1998prototypes} and fitted the prototype- and exemplar-based models on the simulated choices as in the original study. 
    We find that ERMI, like humans, becomes increasingly exemplar-based over trial segments ($\hat{\beta}=-0.01 \pm 0.004; z=-2.54,~p<0.01$) whereas choices from MI are explained almost equally well by exemplar-based and prototype-based learning ($\hat{\beta}=-0.002 \pm 0.005; z=-0.47,~p=0.63$) as shown in Figure \ref{fig:experiment2} (b) and (c). While humans are better explained by prototype-based models for the first five blocks, ERMI is already better explained by an exemplar-based model from the second block onwards. Like in the previous study, this again indicates that ERMI is learning the task faster than humans. Nonetheless, these results demonstrate that training on ecological category learning problems is sufficient for developing an exemplar-based strategy category assignment. We provide additional details and results in Appendix \ref{app:experiment2}.
    
    We then evaluated if ERMI also explains human choices better than competing models in this study. To do this, we conducted a model comparison on human data from \citet{devraj2021dynamics} -- a replication of the \citet{smith1998prototypes} study -- following the procedure outlined in Section \ref{sec:experiment1}. 
    Posterior model frequency in Figure \ref{fig:experiment2} (d) suggests that ERMI explains human choices the best most often ($0.32 \pm 0.06$), closely followed by the GCM ($0.24 \pm 0.05$) and the rule-based model ($0.20 \pm 0.05$). 

    We additionally simulated \textsc{Claude-v2} directly on the \citet{devraj2021dynamics} study. We found that \textsc{Claude-v2} is completely exemplar-based after the second block, which significantly differs from the strategy used by humans and ERMI (see Figure \ref{fig:llmsimulationsdevraj2022}). We also found that ERMI (41207.20) offers a better fit to human choices than \textsc{Claude-v2} (42401.42) in terms of BIC. These results once again confirm that while \textsc{Claude-v2} can generate ecologically valid data, it cannot bring to bear its rich set of ecological priors to act in human-like ways.
    
\section{ERMI displays human-like generalization} \label{sec:experiment3}

    Having shown that ERMI learns category structures in a human-like way, we next inspect how it generalizes to stimuli unseen during training and whether it displays generalization patterns similar to people. To this end, we zoomed into the study from \citet{Johansen2002-xe}, in which participants were instructed to categorize nine four-dimensional stimuli into two categories. The authors then examined how participants generalized to seven transfer stimuli (labeled T1-T7) for which they did not receive feedback during training. In Figure \ref{fig:experiment3} (a), we report the mean probability of participants assigning category A to each of the seven transfer stimuli at the end of the experiment. It can be seen that they assigned stimuli T5, T6, and T7 mostly to category A and T1, T2, T3, and T4 mostly to category B. We provide further details about the paradigm in Appendix \ref{app:experiment3}.

   We simulate the behavior of ERMI and MI on the \citet{Johansen2002-xe} study and found that ERMI generalizes to unseen stimuli in a human-like way by classifying stimuli T1, T3, and T4 more often as category B and T5, T6, and T7 more often as category A. MI, on the other hand, classified all stimuli except T7 mostly as category B. The Euclidean distance between the choice probabilities of humans and MI ($0.67$) was higher than that between humans and ERMI ($0.29$). The pattern of generalization of ERMI matches humans except for stimulus T2, which is classified at around chance level. Why exactly this is the case remains a question for future work. One possible explanation could relate to the observation that T2 only contains two non-zero features (see Figure \ref{fig:experiment3} (d)) while all other stimuli categorized as B contain three non-zero features.


\section{ERMI achieves state-of-the-art performance on machine learning benchmarks} \label{sec:benchmarking}

    Humans can bring to bear the rich set of priors they have acquired from their everyday interactions to generalize to novel tasks \cite{tenenbaum2001generalization, griffiths2006optimal, lake2023human}. Thus far, we have demonstrated how ERMI captures some of these adaptive priors and explains essential aspects of human category learning. This led us to ask whether such a model can also perform well on real-world classification tasks from the machine learning literature.
    
    To investigate this, we evaluated the performance of ERMI on a set of real-world classification tasks from the OpenML-CC18 benchmarking suite \cite{oml-benchmarking-suites}.
    We excluded classification tasks with more than two classes, over 100 features, or missing values, resulting in a set of 23 classification tasks. 
    We then compared the performance of ERMI against several baseline models, including logistic regression, a support vector machine (SVM; \citealp{cortes1995support}), XGBoost \cite{chen2016xgboost}, and TabPFN \cite{hollmann2023tabpfn}.
    TabPFN is an off-the-shelf PFN-based model designed for tabular data prediction that has recently shown state-of-the-art performance on an independent large-scale evaluation \cite{mcelfresh2023neural}.
    
    Following the procedure of \citet{Muller2021-ol}, we created 20 class-balanced learning problems with 100 data points for each of the selected data sets. We provided 30 input-label pairs to our models for training and evaluated them on the remaining 70 data points. For each data set, we reduced the input dimensionality to four, keeping only the features with the highest F-value to the target variable. We measured the performance on the test set based on two metrics: accuracy and rank.
    
    We found that ERMI is the best model in terms of both mean accuracy ($70.95\% \pm 0.54$) and mean rank ($2.26 \pm 0.22$; see Table \ref{tab:performance_metrics}).
    TabPFN is the second-best model in terms of mean accuracy ($70.51\% \pm 0.63$), while XGBoost is the second-best in terms of mean rank ($2.61 \pm 0.30$). We provide the summary of main results in Table \ref{tab:performance_summary} and detailed results across all data sets in Appendix \ref{app:benchmark}.
    
    The performance gain of ERMI over TabPFN in terms of mean accuracy is in the same range as TabPFN over XGBoost, indicating that the improvement is substantial. In terms of parameters, TabPFN has $64$ times more parameters than ERMI, and on disk, it is around $80$ times larger, suggesting that there is room for further improvement. When compared to a parameter-matched PFN and MI, ERMI shows a significant accuracy boost of $3.5\%$ and $10\%$ respectively. 

    We additionally adapted a Bayesian analysis from \cite{stephan2009bayesian} to compute the probability that a model offers the best performance (within a set of alternative models) most frequently across data sets. This measurement is also known as the exceedance probability (EXP) in the statistics literature and is reported in the last row of Table \ref{tab:performance_summary}. We found that ERMI is the best model according to this metric, with an exceedance probability of 0.66. TabPFN comes second with an exceedance probability of 0.24.
    
    \begin{table}[!t]
    \centering
    \caption{Performance metrics on OpenML-CC18 benchmark.}
    \label{tab:performance_summary}
    \vskip 0.14in
        \begin{small}
            \begin{sc}
                \begin{tabular}{lcccccr}
                \toprule
                \textbf{Mean} & \textbf{SVM} & \textbf{XGBoost} & \textbf{TabPFN} & \textbf{ERMI} \\
                \midrule
                \textbf{Acc.} & $69.29\% $ & 70.17\% & $70.51\%$ & \textbf{70.95\%}\\  \midrule
                \textbf{rank} & $2.76$ & $2.61$ &  $2.85$ & \textbf{2.26}\\\midrule
                \textbf{Exp} & $0.01$  & $0.00$ & $0.24$ & \textbf{0.66}\\  
                \bottomrule
                \end{tabular}
            \end{sc}
        \end{small}
    \vskip -0.1in
    \end{table}
    
\section{Discussion}

    Ecological rationality has a long history in cognitive science \cite{brunswik1955representative, simon1990invariants, Todd2012-ta}. The proposition that people are to some degree adapted to the problems they have to solve in the real world is maybe trivial. Figuring out how strong this adaptation is, on the other hand, has remained a big open question. Yet, it has been notoriously difficult to build models that are optimally adapted to the problems that people encounter in their everyday environment. From a technical perspective, the framework of meta-learning offers a solution to this problem. Yet, it has thus far only been applied to artificially-generated environments \cite{kumar2020meta, binz2022exploration, binz2022heuristics, lake2023human, jagadish2023zero}. The main obstacle up to now was that the number of available real-world data sets was insufficient for meta-learning. We have shown that one can overcome this obstacle by prompting LLMs to generate a large collection of ecologically valid category learning tasks. We have then used meta-learning to obtain models that are optimally adapted to them, leading to a class of ecologically rational models that we call ERMI. 
    
    ERMI captured three patterns, which are also observed when humans learn to categorize objects: (1) it showed similar learning difficulties as humans, (2) it became more exemplar-based as learning progressed, and (3) it displayed human-like generalization patterns. Furthermore, it explained human behavior better than competing approaches on a quantitative level, thereby suggesting that we can explain many characteristics of human category learning using the principle of ecological rationality.
    
    The methodology developed, more importantly, enables us to test whether people are ecologically rational or not, thereby allowing us to ask questions such as: how much of human learning can be attributed to data distributional properties alone? The approach we have proposed is quite general and in future work, we plan to extend it to other domains, such as decision-making \cite{bourgin2019cognitive, peterson2021using}, reinforcement learning \cite{brandle2021exploration}, and function learning \cite{schulz2017compositional, schulz2016probing}. Furthermore, it would be interesting to not impose a predefined task structure on the LLM but instead let it synthesize arbitrary task structures by itself.

    While it is possible to come up with a mathematical formulation that allows generating features that match a given statistic like correlation or linearity, it is unclear how to design an algorithm that generates category-learning tasks that match all real-world statistics measured and unmeasured. This is exactly where the strength of pre-trained LLMs lies. If queried correctly, LLMs can act as a single, easy-to-access, rich -- possibly infinite -- source of real-world data.
    
    \section{Limitations}
    
    There are some limitations to generating ecological data that are worth noting. Firstly, it is uncertain whether the current approach can be scaled to generate several more orders of features and multi-class category learning problems. Secondly, strict data curation protocols are necessary if the approach is to be used in cases where unbiased factually correct data are required. Thirdly, preliminary analyses revealed that the data generated by LLMs are US/western-centric \cite{Anthropic2023-hr, tamkin2023evaluating}. Therefore, culturally diverse LLMs are needed to generate data that capture the rich diversity of the real world. 
    
    On the cognitive science front, there were also facets of human category learning not captured by ERMI.  In particular, we found that ERMI generally learned faster than people. In the third experiment, for instance, ERMI already displayed the generalization patterns shown in Figure \ref{fig:experiment3} after two blocks, while people required 16 blocks or more \cite{Johansen2002-xe}. We believe that this gap can (at least partially) be closed by incorporating limited computational resources \cite{binz2022heuristics, jagadish2023zero} or other architectural constraints \cite{achterberg2023building}.
    

    \section{Conclusion} 
    
    We have shown that LLMs can generate ecologically valid category learning tasks that can be used for meta-learning. With these models at hand, we then demonstrated that one can explain human category learning to a large extent using the principle of ecological rationality. Furthermore, the priors acquired by ERMI are rich enough that it achieves state-of-the-art performance on a real-world classification benchmark. In future work, we plan to scale up ERMI's architecture, train it on classification tasks with a flexible number of features, increase the maximum number of data points, and allow for more than two classes.
    

    
    
    \section*{Acknowledgements}
    We thank all reviewers for their constructive and thoughtful feedback. We would also like to thank the authors of \citet{devraj2021dynamics}, \citet{Nosofsky1994-hw}, and \citet{Badham2017-hc} for making the data from their study available. Furthermore, we thank
    the members of the ``Computational Principles of Intelligence Laboratory'' (CPI Lab), participants of the ``Analytical Connectionism'' summer school, Dan John, Yashas Annadani, and Laura Heidiri for their comments, discussions, and support. This work was supported by the Max Planck Society, the Volkswagen Foundation, and funded by the Deutsche Forschungsgemeinschaft (DFG, German Research Foundation) under Germany’s Excellence Strategy–EXC2064/1–390727645.15/18.

    
    
    \section*{Impact Statement}

    This paper presents work whose goal is to show that methods from machine learning, specifically large language models and meta-learning, can be used to advance our understanding of human cognition. An important finding from our work that might be of significance to a broader audience is that training on a large corpus of human-generated text has allowed large language models to capture certain key ecological features. Therefore, one has to be aware that such very powerful features can emerge in foundation models when scaled appropriately. There may be other potential societal consequences of our work, but we do not feel that they need to be specifically highlighted here. 
    

\bibliography{main}
\bibliographystyle{icml2024}

\newpage
\appendix
\onecolumn
\section{Generating ecologically valid category learning tasks using LLMs} \label{app:generate_tasks}

   \subsection{Synthesizing task features and labels} \label{app:synthesize}
   
    We synthesized task features and labels from \textsc{claude-v2} using the prompt mentioned in Section \ref{prompt:task_label}, running it for a total of 100 batches. In each batch, we generate 250 tasks or until a maximum token length of $10$k is reached. We repeat the procedure for all three different stimuli dimensions. In total, we synthesized 23421, 20690, and 13693 category learning tasks with three, four, and six-dimensional features respectively.

\begin{figure*}[!h]
     \centering
     \includegraphics[width=\textwidth]{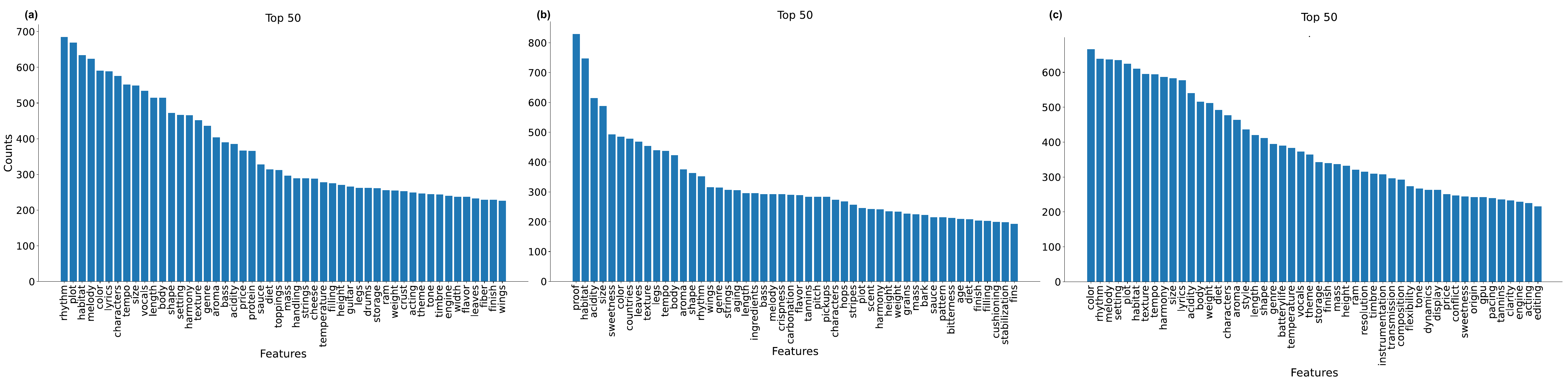}
      \caption{\textbf{Frequency of different features in \textsc{claude-v2}  synthesized category learning tasks:} Counts for the top-50 most frequently occurring task features in the 23421, 20690, and 13693 category learning tasks generated for three (a), four (b), and six-dimensional features respectively.}
     \label{supp:LLMfeatures}
     \end{figure*}

\begin{figure*}[!h]
     \centering
     \includegraphics[width=\textwidth]{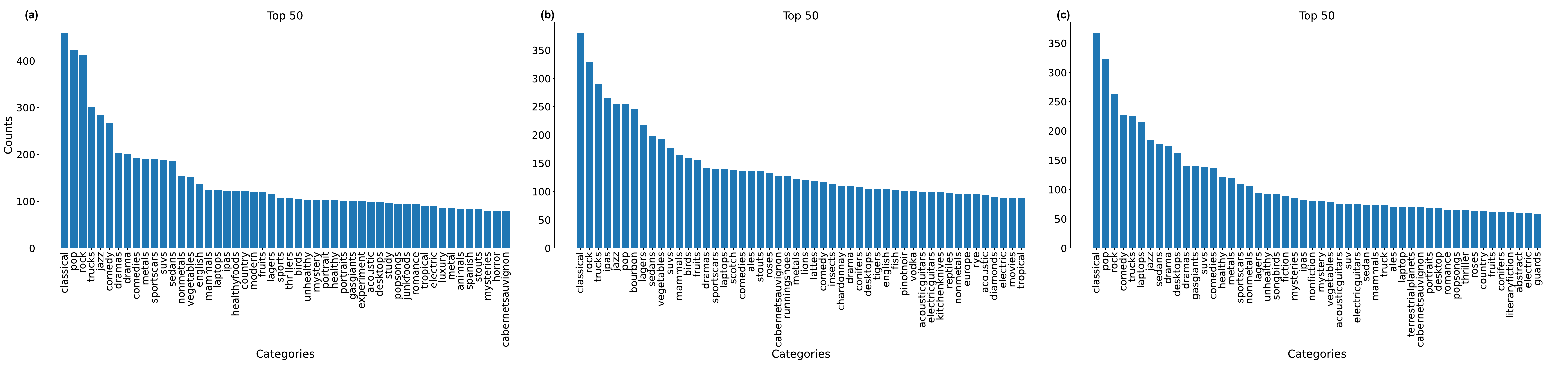}
      \caption{\textbf{Frequency of different categories in  \textsc{claude-v2}  synthesized category learning tasks:}  Counts for the top-50 most frequently occurring category labels in the 23421, 20690, and 13693 category learning tasks generated for three (a), four (b), and six-dimensional features respectively.}
     \label{supp:LLMcategories}
     \end{figure*}

     We show the counts for the top-50 most frequently occurring task features in Figure \ref{supp:LLMfeatures} and categories in Figure \ref{supp:LLMcategories} for the 23421, 20690, and 13693 category learning tasks generated with three (a), four (b), and six-dimensional features respectively. We found that the model tends to produce features belonging to topics such as musicality (for instance, rhythm, melody, lyrics, tempo, vocals) and food (for instance, aroma, texture, crust, diet, protein).  Regarding categories, there were many related to music (for example, classical, pop, jazz, rock) and vehicles (like trucks, SUVs, sedans). In future work, we plan to do a semantic analysis of the generated task features and category labels using methods such as hierarchical clustering to study the semantic grouping of the generated task features/categories. 
     
    \subsection{Generating category learning tasks} \label{app:prompting}
    We used the prompt mentioned in Section \ref{prompt:task_values} to generate data points for task features and labels synthesized in the first stage. We aimed to generate 100, 300, and 616 data points for category learning tasks with stimuli of three, four, and six dimensions respectively. However, sometimes the upper bound of $100$k tokens is reached and the model does not generate the number of data points specified in the prompt. This was especially the case in category learning tasks with higher dimensional stimuli. In these cases, we generate the data points in two steps. In step one, we do the data generation as before using the original prompt. In step two, we query the model again but now conditioned it on the first 20-40 percent of the data points generated in step 1 along with the prompt. This way, we could scale up the generated data points to up around 1.5 times the original length keeping the same underlying data distribution. We generated a total of 11518, 8950, and 12911 category learning tasks with three, four, and six-dimensional stimuli respectively. 

    \paragraph{Note on other LLMs:} We ran preliminary tests on LLMs other than \textsc{claude-v2} for generating category learning tasks including LLaMA \cite{touvron2023llama} and GPT-4 \cite{achiam2023gpt}. The non-instruction-tuned LLaMA version was not able to consistently produce the 100-616 data points we need per task. It was also especially difficult to parse the output of the model as the generated output failed to stick to the provided format. This problem could be mitigated with instruction-tuned versions of LLaMA that are now available but we leave it for future work. 
    We also performed preliminary tests on the GPT-4 model from OpenAI for category learning task generation but found the model sampled the values for the features from uniform distribution using the code module and applied simple heuristic rules most of the time. For example, the sum of two features should be greater than the third and the mean of the two features greater than the other. Upon conducting a preliminary statistical analysis on a relatively small data set generated from GPT-4, we found that task statistics are similar to the statistics of the MI data set, thereby lacking the diversity in terms of measures reported in Section \ref{sec:llm_stats}.

    \subsection{Parsing data generated by LLMs} \label{app:parsing}

    \paragraph{Parsing synthesized task features and labels:} We queried the \textsc{claude-v2} to synthesize task features and labels in the following format: \textsc{feature dimension 1, feature dimension 2, ..., feature dimension {N}, category label 1, category label 2}. 
    We then used a regular expression (regex) pattern, specifically  \verb|\d+\.(.+?)\n|, to efficiently parse and extract relevant data from the model output.  This regex was designed to identify and isolate sequences beginning with a number followed by a period, capturing subsequent characters up to the first newline character. The extracted text was then processed to acquire the names for feature dimensions and category labels by splitting the string at the commas. The final processed data is then stored as a dataframe for future use. 
    
    \paragraph{Parsing generated task data points:} We queried the \textsc{claude-v2} to generate data points for a given category learning task in the following format: \textsc{- feature value 1, feature value 2,..., feature value {N}, category label}. The model followed the aforementioned format while generating the data points for a category learning task, more often than not.  We then used a collection of regular expression (regex) patterns to parse the generated, ensuring accurate handling of different data formats. These regex patterns were designed to cover a wide range of scenarios: capturing numeric values with or without decimal points, handling alphanumeric strings including those with hyphens, and handling complex cases involving commas, hyphens, and optional preceding labels. Furthermore, they extend to different delimiters and formats, from simple comma-separated values to more complex structures with optional components.  In Table \ref{regex-table}, we show all the regex expressions used to parse data points. Based on these expressions, we were able to successfully parse up to 95 \% of the tasks generated by the model. The values extracted from these regex expressions are then stored in a dataframe which acts as an offline repository of tasks on which one can train the ecologically rational meta-learned inference model.

    \begin{table}[!h]
    \caption{Regular expression patterns used for parsing the data points generated for category learning tasks by \textsc{claude-v2}}
    \label{regex-table}
    \vskip 0.15in
    \begin{center}
    \begin{small}
    \begin{sc}
    \begin{tabular}{ll}
    \toprule
    Index & Regular expression \\
    \midrule
    1 & \verb|([\d.]+),([\d.]+),([\d.]+),([\w]+)| \\
    2 & \verb|([\w\-]+),([\w\-]+),([\w\-]+),([\w]+)| \\
    3 & \verb|([-\w\d,.]+),([-\w\d,.]+),([-\w\d,.]+),([-\w\d,.]+)| \\
    4 & \verb|([^,]+),([^,]+),([^,]+),([^,]+)| \\
    5 & \verb|([^,\n]+),([^,\n]+),([^,\n]+),([^,\n]+)| \\
    6 & \verb|(?:.*?:)?([^,-]+),([^,-]+),([^,-]+),([^,-]+)| \\
    7 & \verb|([^,-]+),([^,-]+),([^,-]+),([^,-]+)| \\
    8 & \verb|r'^(\d+):([\d.]+),([\d.]+),([\d.]+),([\d.]+),([\w]+)'| \\
    9 & \verb|r'^(\d+):([\w\-]+),([\w\-]+),([\w\-]+),([\w\-]+),([\w]+)'| \\
    10 & \verb|r'^(\d+):([-\w\d,.]+),([-\w\d,.]+),([-\w\d,.]+),([-\w\d,.]+),([-\w\d,.]+)'| \\
    11 & \verb|r'^(\d+):([^,]+),([^,]+),([^,]+),([^,]+),([^,]+)'| \\
    12 & \verb|r'^(\d+):([^,\n]+),([^,\n]+),([^,\n]+),([^,\n]+),([^,\n]+)'| \\
    13 & \verb|r'^(\d+):(?:.*?:)?([^,-]+),([^,-]+),([^,-]+),([^,-]+),([^,-]+)'| \\
    14 & \verb|r'^(\d+):([^,-]+),([^,-]+),([^,-]+),([^,-]+),([^,-]+)'| \\
    15 & \verb|^(\d+):([\d.]+),([\d.]+),([\d.]+),([\d.]+),([\d.]+),([\d.]+),([\w]+)|\\
    16 & \verb|^(\d+):([\w-]+),([\w-]+),([\w-]+),([\w-]+),([\w-]+),([\w-]+),([\w]+)|\\
    17 & \verb|(\d+):([^,]+),([^,]+),([^,]+),([^,]+),([^,]+),([^,]+),([^,]+)|\\
    18 & \verb|(\d+):([^,\n]+),([^,\n]+),([^,\n]+),([^,\n]+),([^,\n]+),([^,\n]+),([^,\n]+)|\\
    19 & \verb|(\d+):(?:.*?:)?([^,-]+),([^,-]+),([^,-]+),([^,-]+),([^,-]+),([^,-]+),([^,-]+)|\\
    20 & \verb|(\d+):([^,-]+),([^,-]+),([^,-]+),([^,-]+),([^,-]+),([^,-]+),([^,-]+)|\\

    \bottomrule
    \end{tabular}
    \end{sc}
    \end{small}
    \end{center}
    \vskip -0.1in
    \end{table}

\section{Meta-learned inference models}

    \subsection{Synthetic data generation } \label{app:synthetic_data}

    \paragraph{Bayesian logistic regression prior used for training MI model:}
    We generated 10k synthetic binary classification tasks with a linear decision boundary using a Bayesian logistic regression model. To do this, we sample the input features from a normal distribution with zero mean and unit variance for a given number of data points and stimulus dimensions. We then applied a linear transformation, followed by a sigmoid function, and rounded the result to determine the binary class for the given input. The parameters of the linear transformation are sampled from a normal distribution with zero mean and unit variance. The maximum number of data points within a task was set to 400, 650, or 300 for category learning tasks with three, four, and six-dimensional stimuli respectively. These values were chosen depending on the length of the experiments on which these models were evaluated.

    \paragraph{Bayesian neural network prior used for training PFN model:}
    We generated 10k synthetic binary classification tasks using a version of the Bayesian neural network (BNN) prior developed by \citet{Muller2021-ol}. We used normally-distributed i.i.d. input features for a given number of data points and stimulus dimensions. We then pass the input through a BNN with two layers with tanh non-linearity and hidden dimensionality of 64. The network weights and biases were sampled from a normal distribution with a mean of zero and standard deviation of 0.1 and subjected to an additional sparsity constraint (i.e., 20 percent of randomly chosen network weights and biases set to zero). The maximum number of data points was once again set to 400, 650, or 300 for category learning tasks with three, four, and six-dimensional stimuli respectively. The model output is passed through a sigmoid function to generate probability estimates which are then rounded to determine the class for the given input. 
    
    \subsection{Data pre-processing, model architecture, and training} \label{app:model_training}

    \paragraph{Data pre-processing:} We filter out all tasks with more than two unique category labels and then binarize the category labels which are originally strings to make them consistent across tasks. The assignment of category labels, that is either ‘0’ or ‘1’, within a category learning task was randomized during batch creation. This ensures that there can be no unintended correlations between the stimuli seen during training and the labels (across all training data each input vector is assigned half of the time to label ‘0’ and half of the time to label ‘1’). We also normalized each feature independently using a min-max normalization scheme such that values taken by any feature lie always between zero and one. Both the task features and data points were shuffled while generating tasks. Note that the tasks generated by LLMs are typically of different lengths.  Whenever the sampled tasks are of variable lengths, they are padded with zeros to match the length of the longest task sample within the batch. We additionally also sampled LLM-generated data points with replacement to match the length of the experimental task used in the \citet{devraj2021dynamics} and \citet{Johansen2002-xe} studies. We resorted to this strategy as the LLM-generated tasks had a maximum of about 200 data points per task and by resampling, we can evaluate the model on experiments with larger horizons without any drop in performance. The batch size was set to 64 for three- and four-dimensional stimuli and to 32 for six-dimensional stimuli and it operated under a maximum steps regime of 400, 300, and 650 for three, four, and six-dimensional tasks respectively. 
    
    \paragraph{Model architecture and training:} The task features were mapped onto a 64-dimensional embedding space and positional encoded  using sine and cosine functions of different frequencies as in \citet{vaswani2017attention}. A causal attention mask was then generated for the inputs such that the model makes conditional predictions on all preceding data points. The inputs along the attention mask are then passed to the transformer decoder model which had six layers, a model dimension of 64, 256 hidden units in the feed-forward network, and eight attention heads. The output of the transformer was then passed through a linear readout and sigmoid function to generate probability estimates for category 1. In practice, inference for all time steps is performed in parallel by passing a causal attention mask to the \textsc{TransformerDecoder} module in PyTorch \cite{NEURIPS2019_9015}. We used binary cross-entropy (BCE) loss for a given batch of inputs and updated the model parameters using the \textsc{ADAM} optimizer \cite{kingma2014adam} with a learning rate of $10^{-4}$. We trained all our models for a total of \num{500000} episodes. 

\section{LLMs generate ecologically valid category learning tasks} \label{app:real_stats}

\paragraph{Sparsity:} We fitted a logistic regression model for each task and analyzed the sparsity of the resulting regression weights $\mathbf{w} \in \mathbb{R}^d$ using the Gini coefficient $G$:
\begin{equation}
    G(\mathbf{w}) = \dfrac{\displaystyle\sum\limits_{i=1}^d \displaystyle\sum\limits_{j=1}^d |\mathbf{w}_i - \mathbf{w}_j|}{2 d \displaystyle\sum\limits_{i=1}^d \mathbf{w}_i}
\end{equation}

\paragraph{Linearity:} We fitted a logistic regression model and a logistic regression with second-order polynomial features on data $\mathcal{D}$ from each task. We then computed the Bayesian information criterion (BIC) for both models and used them to approximate the posterior
probability that the linear model offers a better explanation of the data (assuming a uniform prior over models):
\begin{align}
    p(M = {\text{linear}} | \mathcal{D}) \approx& \frac{\exp(-0.5 \cdot \text{BIC}_\text{linear})}{\sum_{m \in \{\text{linear, polynomial} \} } \exp(-0.5 \cdot \text{BIC}_m)}
\end{align}

\begin{figure*}[!b]
\centering
\includegraphics[width=\textwidth]{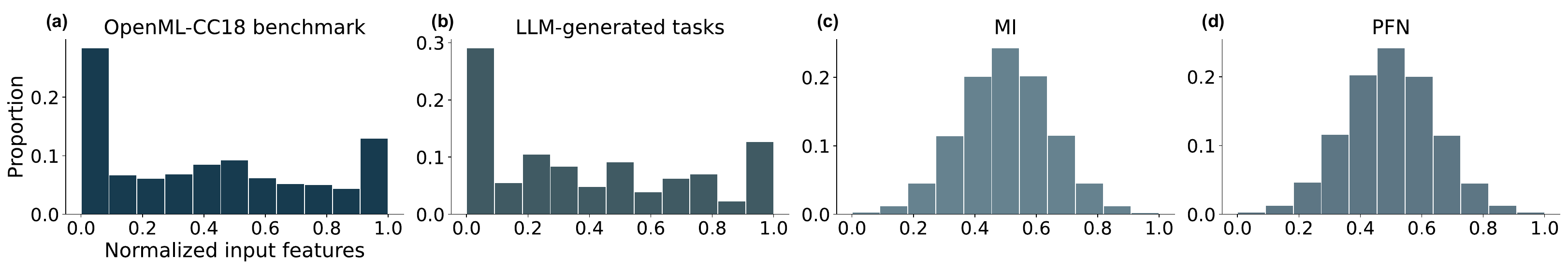}
\caption{\textbf{Comparing input features of real-world classification tasks and LLM-generated tasks}: (a) Histogram of normalized input features from 28 different real-world binary classification tasks from the OpenML-CC18 benchmarking suite, (b) from ecologically valid category learning tasks generated from \textsc{Claude-v2}, (c) synthetic category learning tasks derived using the Bayesian logistic regression prior that were used to train the meta-learned inference (MI) model and (d) synthetic category learning tasks with nonlinear decision boundary derived using the Bayesian neural network prior that were used to train prior-fitted networks (PFN) model.}
\label{fig:compare_inputfeatures}
\end{figure*}

\paragraph{Features values:} We compared the distributions of input values generated by LLMs to input values from real-world datasets and found them to be extremely similar, as shown in Figure \ref{fig:compare_inputfeatures}. We furthermore checked if there were spurious values in the LLM-generated data such as peaks at multiples of 5s and 10s, and found them to be at chance level (20\% for multiples of 5s, 10\% for multiples of 10s).

\begin{figure*}
    \centering
    \includegraphics[width=\textwidth]{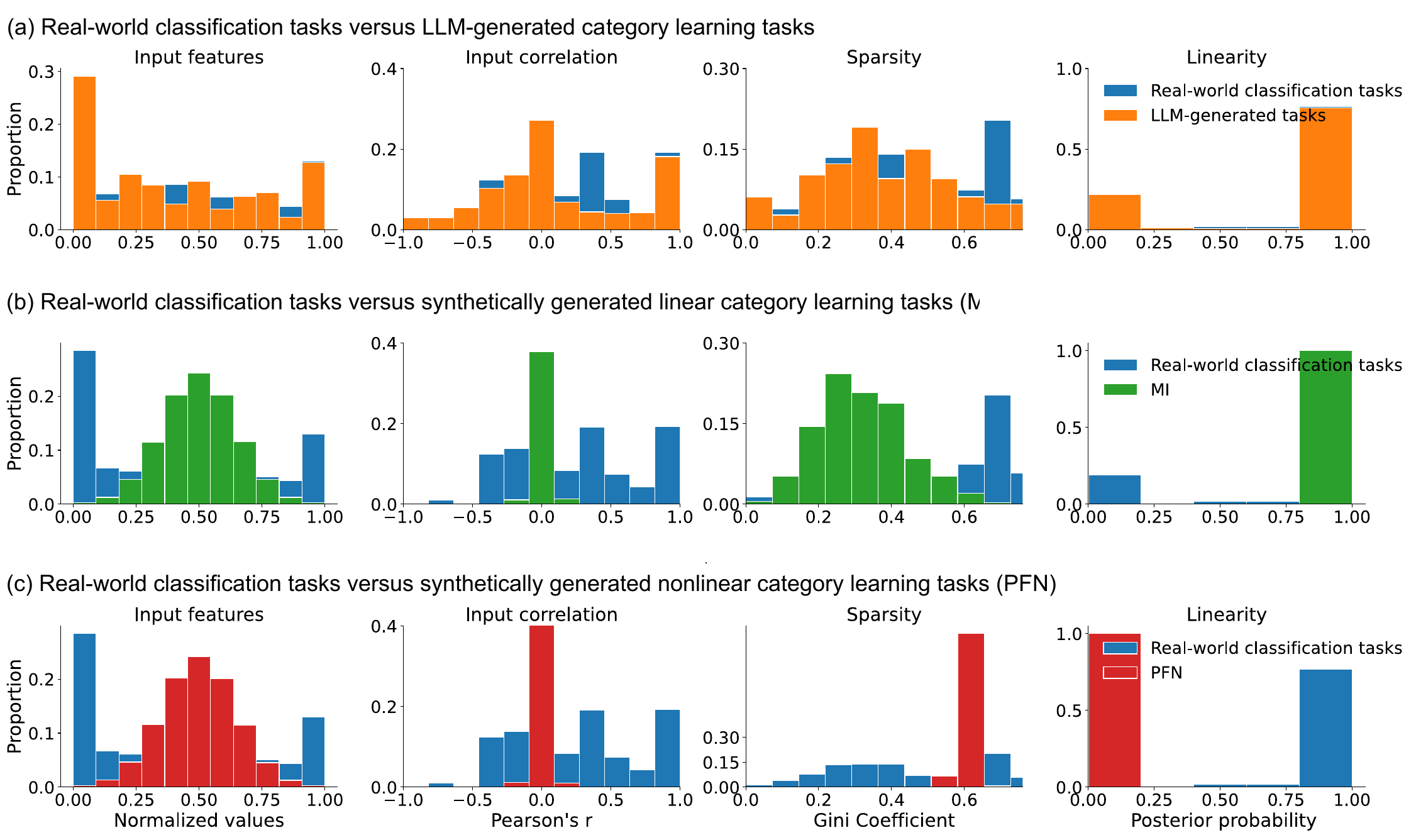}
    \begin{center}
    \caption{\textbf{Comparing the ecological validity of different category learning tasks:} Histogram of normalized input features (first column). Histogram of Pearson's correlation coefficients computed between pairs of features (second column). Histogram of Gini coefficients computed over the logistic regression weights (third column). Linearity of the category learning task (fourth column). These four statistics were (a) ecologically valid category learning tasks generated from \textsc{Claude-v2}, (b) synthetic category learning tasks derived using the Bayesian logistic regression prior that were used to train the meta-learned inference (MI) model, and (c) synthetic category learning tasks with nonlinear decision boundary derived using the Bayesian neural network prior that were used to train prior-fitted networks (PFN) model were overlaid on the same statistics computed on 28 different real-world binary classification tasks from the OpenML-CC18 benchmarking suite.}
    \label{fig:overlaid_stats}
    \end{center}
\end{figure*}

\paragraph{Additional analysis:} We furthermore computed the KL divergence between the histograms of the real-world data set and the synthetically generated data sets for each investigated metric. The resulting KL divergences confirm that the LLM-generated data provides a good proxy for real-world data (see Table 
\ref{tab:datacomparison}). In Figure \ref{fig:overlaid_stats}, we overlaid the histograms of the real-world data set and the synthetically generated data sets for the four different statistics discussed in the main paper. 

\begin{table}[!h]
    \centering
    \caption{KL divergence between real-world data classification data and the synthetically generated data for different statistical measures}
    \label{tab:datacomparison}
    \vskip 0.14in
        \begin{small}
            \begin{sc}
                \begin{tabular}{lcccccr}
                \toprule
                \textbf{Real world tasks vs.} & \textbf{Input Correlation} & \textbf{Gini Coefficient} & \textbf{Posterior Probability} & \textbf{Features} \\
                \midrule
                \textbf{LLM-generated} & $0.2716$	& $0.2286$ & $0.0202$ &	$0.0418$ \\ 
                \midrule
                \textbf{MI Prior} & $7.7637$	& $2.8170$ & $2.5265$ &	$1.6758$ \\ \midrule 			
                \textbf{PFN Prior} & $7.7746$	& $8.6391$ & $2.5265$ &	$1.6775$ \\ 
                \bottomrule
                \end{tabular}
            \end{sc}
        \end{small}
    \vskip -0.1in
    \vspace{-0.20cm}
    \end{table}

\clearpage
\newpage

\section{Cognitive models} \label{app:cogmodels}
    In this section, we will provide details regarding the five cognitive models we used for model comparison. 

    \paragraph{Rational model of categorization (RMC):} 
    The RMC is a Bayesian model of human category learning \cite{Anderson1991-ii}. In this paper, we used a meta-learned version of the model, which was obtained using the following data-generating distribution described in \citet{Badham2017-hc}. Model architecture and training followed the protocol used for ERMI, MI, and PFN. We set the free parameters based on an earlier study \cite{Nosofsky1994-hw} to the following values: $c=0.318$, $s_P=0.488$, and $s_L=0.046$. Note that we did not account for these parameters in our model comparisons, which slightly overestimates the ability of the RMC to explain human behavior.

    \paragraph{Prototype-based model (PM):} There are several versions of the prototype model \cite{Medin1978-xf, smith1998prototypes}. Here, we use the version from \citet{smith1998prototypes}. The prototype model assigns a category to an observed stimulus based on the similarity to the category prototypes. The raw distance between the stimulus and a prototype, $q_k$, for category $k$ is computed as a weighted sum of absolute differences across $n$ feature dimensions with weights, $w_j \in [0, 1]$, for the features contained to sum to 1 as shown in Equation \ref{eq:distance}.
    \begin{equation}
        d_{x, q_k}=\sum_{j=1}^n w_j\left|x_j-q_{k, j}\right|,
        \label{eq:distance}
    \end{equation}
    Note that prototypes themselves can either be learned or provided during model definition. Here, we learn the prototypes for the two categories $\{q_1, q_2\}$. Therefore,  $q_{k,j} \in [0, 1.]~ \forall j = \{1, 2,...n\}$ are themselves also parameters.
    Distance is then converted to psychological similarity between prototypes and stimuli using:
    \begin{equation}
    \eta_{x,q_k} = e^{-c \cdot d_{x,q_k}}
    \label{eq:similarity}
    \end{equation}
    where $c$ is a sensitivity parameter that can shrink or amplify discriminability in psychological space. The probability of the stimulus being assigned to category $1$ is then computed using:
    \begin{equation}
        P(k=1 \mid x) = \frac{\eta_{x,q_1}}{\eta_{x,q_1} + \eta_{x,q_2}}
        \label{eq:prob_assignment}
    \end{equation}

    Furthermore, the final model predicted likelihood is a mixture between the predicted probability from the model and a random guess, with guessing parameter $\epsilon$ controlling the mixture probabilities:
    \begin{equation}
        p(k=1 \mid x) = (1 - \epsilon) P(k=1 \mid x) + \epsilon \cdot \mathrm{K}^{-1}
        \label{eq:gcm_pm}
    \end{equation}
    where $\mathrm{K}$ indicates number of categories.
    
    \paragraph{Generalized context model (GCM):} We used the GCM developed by \citet{nosofsky1986attention}. The GCM categorizes an observed stimulus to a category by comparing the sum of its similarity to all previously seen exemplars in each category, $\{C_1, C_2\}$. The raw distance and similarity between observed stimulus and exemplars were computed based on Equations \ref{eq:distance} and \ref{eq:similarity} respectively. The probability of assigning a stimulus to category $k=1$ is then computed based on the summed category similarities in the following way:
    \begin{equation}
        P(k=1 \mid x) = \frac{\sum_{y \in C_1} \eta_{x,y}}{\sum_{y \in C_1} \eta_{x,y} + \sum_{y \in C_2} \eta_{x,y}}
    \end{equation}

    The final model predicted category probabilities is again a mixture between the predicted probability from the model and a random guess as mentioned in Equation \ref{eq:gcm_pm}.

     \paragraph{Rule (Rule):} The rule model implemented in this work assigns a category based on one of the two rules, whichever explains the participant data better. The first rule is based on the values taken by stimulus features along one dimension, and the second is based on the application of the conjunctive rule on pairs of features -- whether a given pair of stimulus features take on the same value. The final model predicted category probabilities is again a mixture between the category prediction from the model and a random guess as mentioned in Equation \ref{eq:gcm_pm}.
    
    \paragraph{Rule plus exception model (Rulex):} We use the same implementation as the Rule model but provide exceptions as inputs to the model. For \citet{devraj2021dynamics} task, we provide $[1, 1, 1, 1, 0, 1]$ as the exception stimulus for category 1, and for category 2, it was set to $[0, 0, 0, 1, 0, 0]$. For \citet{Badham2017-hc} task, we provide $[1, 1, 1]$  and  $[0, 0, 0]$ as exceptions for \textsc{type-2} task. The final model predicted category probabilities is again a mixture between the category prediction from the model and a random guess as mentioned in Equation \ref{eq:gcm_pm}. In future work, we plan to implement a more detailed version of rule-plus-exception model from \cite{Nosofsky1994-gu} where the model learns the exceptions along with the rule. 

    \paragraph{\textsc{Claude-v2}:} We used \textsc{Claude-v2} as a cognitive model of human category learning. To do that, we prompted \textsc{Claude-v2} by paraphrasing the instructions given to participants, see Appendix \ref{app:experiment1} and \ref{app:experiment2} for task-specific prompts. In essence, the model was instructed via the prompt that it would be presented with stimuli belonging to two categories and would receive feedback regarding the correct category following each stimulus. It was instructed to learn a rule based on stimulus features to assign the stimulus to the correct category with increasing experience. As the Claude API only returns the \textsc{Claude-v2} output token and not its probability, the model's prediction was coded as a binary variable, $\pi(k=1 \mid x_t; x_{1:t-1}, y_{1:t-1})$. The final model predicted category probabilities is again a mixture between the category prediction from the model and a random guess as mentioned in Equation \ref{eq:llm}.

     \begin{equation}
        p(k=1 \mid x_t) = (1 - \epsilon) \pi(k=1 \mid x_t; x_{1:t-1}, y_{1:t-1}) + \epsilon \cdot \mathrm{K}^{-1}
        \label{eq:llm}
    \end{equation}

\section{Fitting models to human data} \label{app:fitting}
    In this section, we explain the fitting procedure used to fit the parameters of the models to human data. The model parameters were fit to the data using maximum likelihood estimation. We explain the implementation details for the different model classes below.  The full list of fitted parameters for each model is shown in Table \ref{table:model_params}. 
    
    \paragraph{MI, PFN, RMC and ERMI:}
    We fit an inverse temperature term $\beta$ within the sigmoid function, which squashes the output from the final layer of the transformer to be within $[0,1]$, to each participant. Note that this term is set to one during the meta-learning phase to derive an optimal model and is fitted only during the evaluation phase with the rest of the model weights being frozen. 
    We use the differential evolution optimizer available in the SciPy optimization library \cite{2020SciPy-NMeth}
    for fitting. 

    \paragraph{GCM and PM:} Both models predict the probability of picking a category in a trial-by-trial fashion conditioned on all preceding stimulus-target pairs. We fit their parameters to human choices that minimize the negative log-likelihood of human choices under the model prediction.  To do so, we used the \textsc{minimize} module available in SciPy's optimization library. As mentioned in Section \ref{app:cogmodels}, the weights for the features were bounded to be within $[0,1]$ and sum to 1, the sensitivity term bounded to be within $[0, 20]$. The prototype model requires learning the prototypical stimulus for each category (same dimensionality as the input stimulus, with the feature values bounded to be within $[0,1]$). Both models learning a guessing parameter $\epsilon$, which was bounded to be within $[0,1]$.

    \paragraph{Rule and Rulex:} We used the same procedure as above except that we learn the stimulus dimension $v_i$ on which the rule is applied. 

    \paragraph{\textsc{Claude-v2}:} We used the same procedure as above except that only the guessing parameter, $\epsilon$, is learned. 
    
    \begin{table}[!h]
    \caption{Fitted parameters in each model where $\beta$ is the inverse temperature term, $w_{i}$ indicates the weights for the stimulus feature dimension $i$, $n$ is the number of stimulus feature dimensions, $c$ is the sensitivity term, $\epsilon$ is noise term in an epsilon greedy policy, $q_1$ and $q_2$ are the values for the prototypes for $d$ stimulus features, and $v_i$ are the stimulus dimension on which the rule is applied.}
    \label{table:model_params}
    \vskip 0.15in
    \begin{center}
    \begin{small}
    \begin{sc}
    \begin{tabular}{lll}
    \toprule
    \textbf{Model}     & \textbf{Parameters} &\\  
    \midrule
    $\text{ERMI, MI, PFN, RMC}$ &  $\beta$ \\ 
    $\text{GCM}$ &   $c, \epsilon, w_{i}$ & $\forall ~i \in\{1,2, \ldots, n\}$ \\
    $\text{PM}$ &  $c, \epsilon, w_{i}, q_{1,i}, q_{2,i}$ & $\forall ~i \in\{1,2, \ldots, n\}$ \\
    $\text{Rule}$ &  $v_1, v_2, \epsilon $ \\ 
    $\text{Rulex}$ &  $v_1, v_2, \epsilon$ \\ 
    $\text{\textsc{Claude-v2}}$ &  $\epsilon$ \\ 
    
    \bottomrule
    \end{tabular}
    \end{sc}
    \end{small}
    \end{center}
    \vskip -0.1in
    \end{table}

\section{Bayesian model comparison} \label{app:comparison}
    
    In this section, we provide details regarding the Bayesian model comparison procedure used to compare the fits of different models to the behavioral data. We first performed maximum likelihood estimation to fit model parameters $\theta_m$. We then computed the Bayesian information criterion (BIC) for model $m$ for a given participant as follows: 
    \begin{equation}
        \text{BIC}_{m} = -2 \cdot \max _{\theta_m} \sum_{t=1}^{T} \log p_{\theta_m} \left(\hat{y_{t}} \mid x_{1: t}, y_{1:t-1} \right) + |\theta_m| \log (T)
    \end{equation}

    where $|\theta_m|$ is the number of parameters estimated for model $m$, $T$ is the number of trials in the task and  $\hat{y_{t}}$ is the category choice made by the participant in trial $t$.   BIC penalizes the model based on its complexity and can be used as a measure for comparing goodness-of-fit when models differ in terms of their number of parameters. 
    
    We reported two metrics in the paper: posterior model frequency (in the main text) and exceedance probability. To compute them, we used a Python implementation of the Variational Bayesian Analysis (VBA) toolbox \cite{daunizeau2014vba}. The toolbox requires us to provide log-evidences for each model and participant, which we approximate using $-0.5 \cdot \text{BIC}_{m}$. For further details about this model comparison procedure, see \citet{rigoux2014bayesian}.

\section{ERMI shows human-like difficulty effects} \label{app:experiment1}
    \subsection{Experiment details for \citet{Shepard1961-yu} and \citet{Nosofsky1994-hw}}

     In their replication of the \citet{Shepard1961-yu} study, \citet{Nosofsky1994-hw} conducted the study on 120 participants. The authors used geometric stimuli that varied in shape (squares or triangles), interior line type (solid or dotted), and size (large or small). Every participant completed two problems, therefore, each problem was performed by 40 participants. The participants were informed that the rules for each problem were independent. Following the methodology of \citet{Shepard1961-yu}, the learning process involved classifying stimuli into two categories and receiving feedback. This process was repeated over several blocks (containing up to 16 trials) with randomized stimulus order in each block. Learning in the task was measured until participants achieved a no-error streak in four consecutive sub-blocks of eight trials or reached a maximum of 400 trials. For more details, please refer to \citet{Nosofsky1994-hw}.

     In tasks belonging to \textsc{type 1}, stimuli were assigned to a category depending on the values they take along one of the three dimensions, whereas in \textsc{type 2} tasks, stimuli were assigned to a category by applying the exclusive-or rule along two relevant dimensions. Category assignment in tasks belonging to \textsc{type 3}, \textsc{type 4}, and \textsc{type 5} used a unidimensional rule-plus-exception structure with some stimuli grouped in the central region and some in the periphery. Lastly, \textsc{type 6} tasks require considering feature values along all dimensions. For the illustration of category structures for the six types, please refer to Figure 1 in \cite{Nosofsky1994-hw}.

     In \citet{Badham2017-hc}, the authors replicated the \citet{Shepard1961-yu} study on 96 adults aged between 18 to 87 years. They used eight geometric shapes varying in size (large or small), shape (square or triangle), and color (black or white) in the experiment with the stimuli shown on a mid-gray background. The order of stimuli and their category assignment were randomized. They only considered the first four types from the  \citet{Shepard1961-yu} study but unlike their study, participants performed all four types. Participants performed each task type for a total of six blocks with each block containing 16 trials (resulting in a total of 96 trials) or until they reached a criterion of perfect performance in two consecutive blocks. For more details, please refer to \citet{Badham2017-hc}.

     \subsection{Simulations:} To run simulations of the \citet{Shepard1961-yu} study on  ERMI, MI, and PFN model, the geometric stimuli used in the experiment as mentioned above are converted into binary coded vectors taking values along the three stimulus feature dimensions. The value assignment for a stimulus feature was randomized in every run, the order of presentation of the stimulus was also randomized, and the number of presentations of a stimulus per block was matched to the original study. In each run, the model was evaluated on a task of one particular type.
     
    \begin{figure*}[!h]
     \centering
     \includegraphics[width=\textwidth]{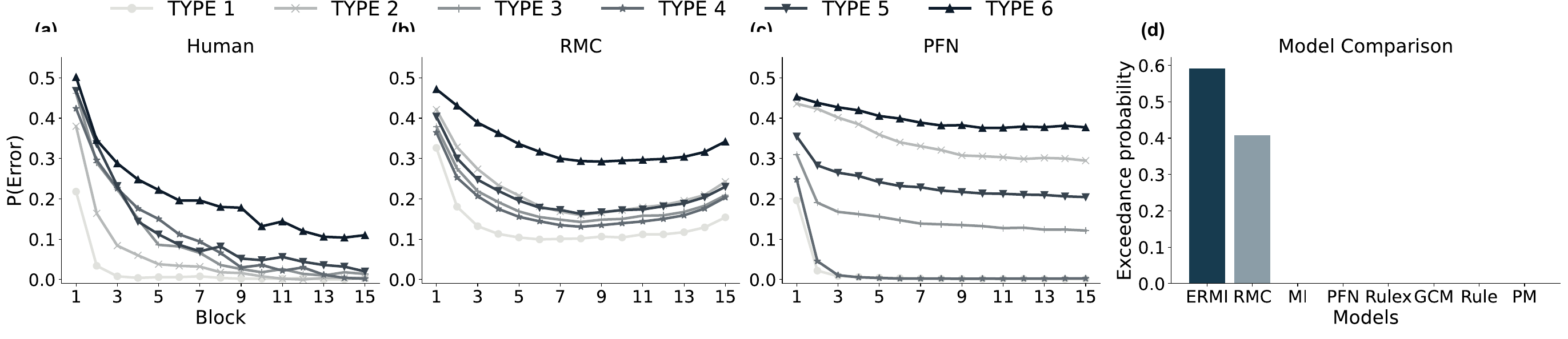}
      \caption{\textbf{Supplementary figure accompanying Figure \ref{fig:experiment1}}: \textbf{(a-c)} Average error probabilities for each task \textsc{type} in each block of 16 trials for (a) humans, (b) RMC, and (c) PFN. \textbf{(d)} The exceedance probability of participants' choices in the \citet{Badham2017-hc} study for eight computational models. Human data in (a) was reproduced from Table 1 in \citet{Nosofsky1994-hw}. RMC and PFN were simulated on \textsc{type 1-6} tasks for 50 runs with the inverse temperature that resulted in the lowest mean-squared error compared to humans, which was $\beta=0.9$ for ERMI, and $\beta=0.9$ for MI.}
     \label{supp:experiment1}
     \end{figure*}
     
    \subsection{Additional observations and results}

     \paragraph{Why are \textsc{type 2} and \textsc{type 6} hard?} We think that this could be because \textsc{type 2} tasks involve applying the exclusive-or rule along two relevant dimensions (and ignoring one of the dimensions altogether), while \textsc{type 6} tasks require memorizing feature values taken by stimuli along all dimensions, making it hard for models to learn.

     \paragraph{Learning curves of RMC and PFN are not as similar to humans as ERMI:} MSE between learning curves of humans and RMC and PFN was $0.10$ and $0.17$ respectively. They are larger than that of ERMI which was $0.03$.
     
     \paragraph{ERMI learns the task faster than people:} We transform the block variable $t$ using an exponential kernel as follows: 
     \begin{equation}
         y = a e^{-b*t} + c
     \end{equation}
     where $a$ is the amplitude, $b$ is the decay coefficient term, and $c$ is the offset term, and then regressed the transformed variable onto the error rate for both ERMI and humans. We found that the fitted decay coefficient for ERMI ($1.24$) is larger than for humans ($0.44$).

     \paragraph{Baseline models fit the data adequately:} Baseline models particularly the GCM $(177.35 \pm 4.70)$ and PM  $(201.19 \pm 3.38)$ model did fit the data quite well in terms of log-likelihoods compared to ERMI $(200.82 \pm 4.56)$. However, the number of parameters being fit to human data is quite large in these models (refer to Table \ref{table:model_params}). Therefore, they are heavily penalized and have higher BIC values than ERMI.

    \begin{table}
    \centering
    \caption{Mean performance of humans and models for each rule type in replication of \citet{Shepard1961-yu} study over 15 blocks. Human data was taken from Table 1 in \citet{Nosofsky1994-hw}. Details of model simulations can be found in Appendix \ref{app:experiment1}.} 
    \vspace{2mm}
    \begin{tabular}{rccccccc}
    \toprule
     \multicolumn{1}{c}{\textbf{Model}} & \multicolumn{6}{c}{\textbf{Rule}}  & \textbf{MSE} \\
     \cmidrule(lr){2-7} 
      \multicolumn{1}{r}{}& \textsc{Type 1} & \textsc{Type 2}  &  \textsc{Type 3} &  \textsc{Type 4} &  \textsc{Type 5} &  \textsc{Type 6}  \\
    \hline
    Humans & .0201 & .0565  & .1015 & .1120 & .1212 & .2048 & .0000 \\
    \hline
    ERMI & .0586 & .0891 & .0855 & .0826 & .0888 & .1172 & \textbf{.0287 } \\
    \hline
    MI & .0686 & .4089 & .2404 & .1431 & .2880 & .4201 & .2627 \\
    \hline
    PFN & .0170  & .3405  & .1533  & .0226  & .2371  & .3975 & .1736  \\
    \hline
    RMC & .1329  & .2215  & .1903  & .1718  & .2132  & .3364 & .1003  \\
    \bottomrule
    \end{tabular}
    \label{table:shepardtask}
    \end{table}
    

    \subsection{\textsc{Claude-v2} as a cognitive model of human category learning}

     \paragraph{Simulations:} To run simulations of the \citet{Badham2017-hc} study on \textsc{Claude-v2}, we queried it using the prompt shown in Appendix \ref{prompt:shepard1961task}. The geometric stimuli used in the experiment are described in textual format. The order of stimulus presentation was randomized, and the number of stimulus presentations per block was matched to the original study. We ran 96 simulation runs for each of the six rules. 

    \begin{tcolorbox}[sharp corners, colback=mBlue!5!white,colframe=mBlue!75!black, width=1.\textwidth, left=4pt,right=4pt, top=4pt, bottom=4pt, title=\textbf{Prompt for  Badham et al. 2017 study} \label{prompt:shepard1961task}]
       In this experiment, you will be shown examples of geometric objects. Each object has three different features: size, color, and shape. Your job is to learn a rule based on the object features that allows you to tell whether each example belongs in the \{A\} or \{B\} category. As you are shown each example, you will be asked to make a category judgment and then you will receive feedback. At first you will have to guess, but you will gain experience as you go along. Try your best to gain mastery of the \{A\} and \{B\} categories. \\
    
        - In trial 1, you picked category \{A\} for Big Black Square and category \{A\} was correct.\\
        - In trial 2, you picked category \{A\} for Small Black Triangle and category \{B\} was correct\\
    
    Human: What category would a Small Black Triangle belong to? (Give the answer in the form ``Category $\langle$your answer$\rangle$").
    
    Assistant: Category
    \end{tcolorbox}

    \begin{figure*}[!h]
    \centering
    \includegraphics[width=\textwidth]{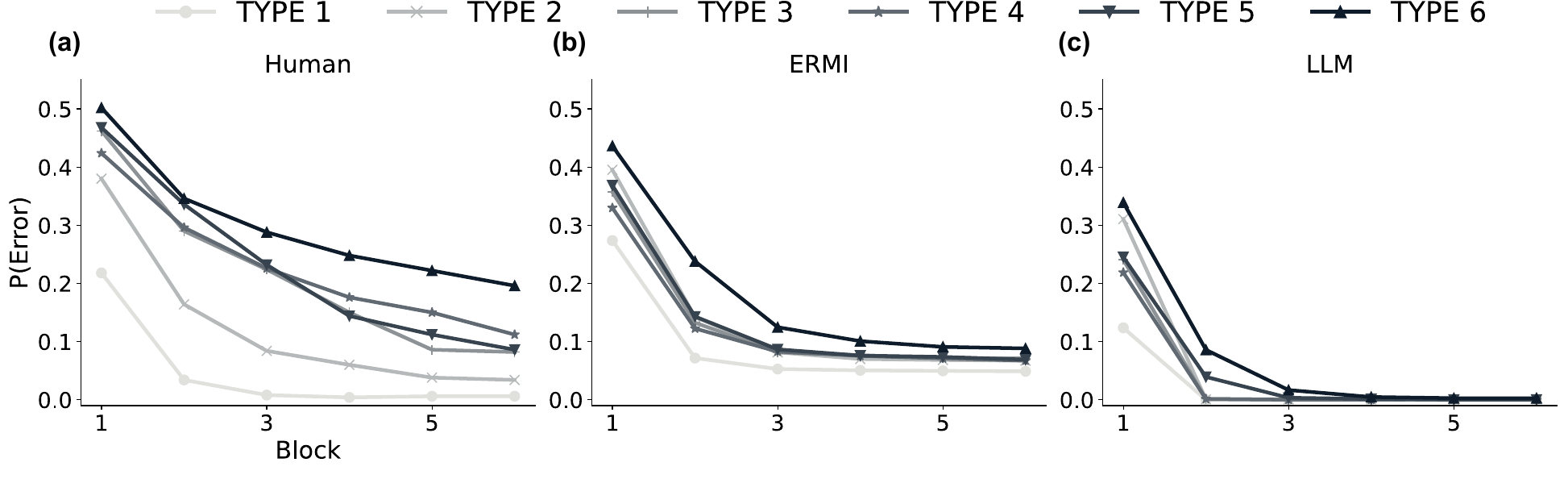}
    \caption{\textbf{Unlike ERMI, \textsc{Claude-v2} does not show human-like learning difficulties}: \textbf{(a-c)} Average error probabilities for each task \textsc{type} in each block of 16 trials for (a) humans, (b) ERMI, and (c) LLM. Human data in (a) was reproduced from Table 1 in \citet{Nosofsky1994-hw}. ERMI was simulated on \textsc{type 1-6} tasks for 50 runs with the inverse temperature set to $\beta=0.4$. \textsc{Claude-v2} was simulated for 94 runs each on \textsc{type 1-6} tasks with temperature term set to 0.}
    \label{fig:llmsimulationsbadham2017}
    \vspace{-0.20cm}
    \end{figure*}

    \begin{table}[!b]
    \centering
    \caption{Mean performance of humans and models for each rule type in the \citet{Badham2017-hc} study. Details of model simulations can be found in Appendix \ref{app:experiment1}.} 
    \vspace{2mm}
    \begin{tabular}{rccccccc}
    \toprule
     \multicolumn{1}{c}{\textbf{Model}} & \multicolumn{6}{c}{\textbf{Rule}}  & \textbf{MSE} \\
     \cmidrule(lr){2-7} 
      \multicolumn{1}{r}{}& \textsc{Type 1} & \textsc{Type 2}  &  \textsc{Type 3} &  \textsc{Type 4} &  \textsc{Type 5} &  \textsc{Type 6}  \\
    \hline
    Human &  .0460 & .1267 & .2157 & .2307  &  .2297 & .3003 & .0000 \\
    \hline
    ERMI & .0912 &  .1374 & .1317 & .1248 & .1361 & .1798  &  \textbf{.0538}\\
    \hline
    LLM &  .0206 & .0521 & .0404 & .0368 & .0482 &  .0752 & .1771 \\
    \hline
    MI & .0959 &  .4337 & .2655 & .1657 & .3175 & .4387 &  .1795 \\
    \hline
    PFN & .0402 &  .3908 &  .1887 & .0527 & .2722 & .4237 & .1541   \\
    \hline
    RMC & .1593 &   .2744 &  .2314 & .2164 & .2576 & .3845 & .0564   \\
    
    \bottomrule
    \end{tabular}
    \label{table:shepardtaskwithllm}
    \end{table}

\section{ERMI becomes more exemplar-based} \label{app:experiment2}
    \subsection{Experiment details for \citet{smith1998prototypes} and \citet{devraj2021dynamics}}
    \citet{smith1998prototypes} conducted their study on 32 participants, using 14 six-dimensional stimuli, with each stimulus mapping to a six-letter nonsensical word such as gafuzi, kafitdo, nivety, wysero, etc. --- see Appendix A of \citet{smith1998prototypes} for all words. Each stimulus can be represented by a six-digit binary string, where each digit and position corresponds to a specific letter. For example, if the stimulus 'gafuzi' corresponds to the binary code '000000', then 'gyfuzi' corresponds to '010000', and so on. The stimuli were assigned to categories such that stimulus '000000' corresponds to category 1 and stimulus '11111' corresponds to category 2. 
    We only considered the non-linearly separable (NLS) category structure from Experiment 2 in this work. According to this, a category contained five stimuli with five features in common with the prototype, and one stimulus with five features in common with the opposing prototype. Therefore, category 1 contained seven stimuli as follows: [000000, 100000, 010000, 001000, 000010, 000001, 111101]. The remaining seven stimuli belonged to category 2 [111111, 011111, 101111, 110111, 111011, 111110, 000100]. Participants had unlimited time to make their choice on each trial. After making their choice, they were told whether it was a correct decision or not. Participants completed a total of 560 trials, or 10 blocks (called trial segments by the authors but we call them blocks to be consistent with other experiments) of 56 trials each, in which they saw each stimulus four times.   

    \citet{devraj2021dynamics} replicated the task as mentioned above and collected data from 60 participants. Participants were recruited from the 18-23 age range and English-speaking population using Prolific. Their study involved 11 blocks instead of 10 and as a result, they had 616 trials.

     \subsection{Model-based analysis:} The 616 choices made by participants and meta-learning models were divided into 11 blocks of 56 trials each. We obtained the choices from the models --  ERMI, MI, and PFN -- by simulating them on the task for a total of 50 runs using the $\beta$s fitted to participants in the \citet{devraj2021dynamics} study. We then fit prototype- and exemplar-based models onto the choices of humans and models to see if they are better explained by prototype or exemplar-based strategy. To fit their parameters, we minimize the sum of squared errors (SSE) between observed and predicted probabilities for each participant for a given block following the original study's approach: 
    \begin{equation}
        SSE = \sum_{t=1}^{14} (p(k=1|x_t) - \hat{p}_{1,x_t})^2
    \end{equation}
    where $p(k=1|x_t)$, from Equation \ref{eq:gcm_pm}, is the predicted probability from the model -- either GCM or PM -- that stimulus $x_t$ belongs to category 1 based on an entire trial segment (56 trials) of data, and $\hat{p}_{1,x_t}$ is the proportion of trials in the trial segment (out of those in which stimulus $x_t$ was seen) in which the participant or model categorized stimulus $x_t$ to category 1. We used SciPy's Sequential Least Squares Programming (SLSQP) method in the SciPy's optimization module to obtain the best fitting parameter for the two models as in \cite{devraj2021dynamics}. We then compared the SSE computed using the best-fitting parameters between the two models as shown in Figure \ref{fig:experiment2}(a).


    \begin{figure*}[!h]
     \centering
     \includegraphics[width=\textwidth]{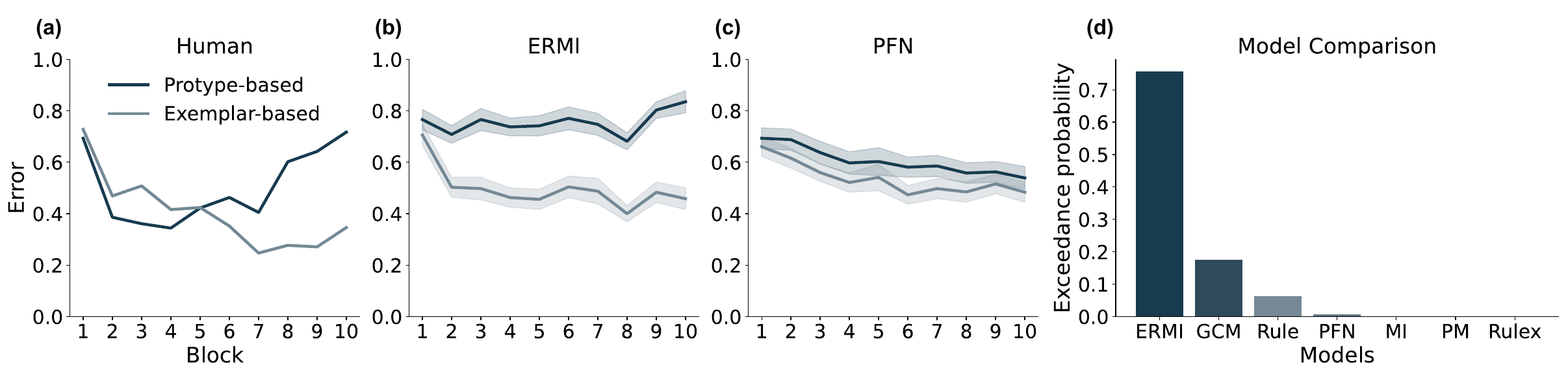}
      \caption{\textbf{Supplementary material accompanying Figure \ref{fig:experiment2}:}\textbf{(a-c)} The average error of exemplar- and prototype-based models fitted to (a) human choices, (b) simulated choices from ERMI, and (c) simulated choices from PFN for each block of 56 trials. \textbf{(d)} The exceedance probability of participants’ choices in the \citet{devraj2021dynamics} study for seven computational models. Human data in (a) was reproduced from \citet{smith1998prototypes}. ERMI and MI were simulated using inverse temperature values fitted to participants' choices in \citet{devraj2021dynamics}. The mean of the fitted inverse temperature and its standard error were $0.09 \pm 0.01$ for ERMI and $0.14 \pm 0.01$ for MI, respectively. The shaded region shows the standard error of the mean.}
     \label{supp:experiment2}
     \end{figure*}

    \subsection{Additional observations and results}

    \paragraph{ERMI is better explained by the exemplar-based model with learning: } We fit a linear mixed-effects model to measure this effect quantitatively. We predict the average error term from the GCM and PM model using blocks and model as predictors and test the interaction between blocks and model. Blocks (1-10) were mean-centered and the exemplar model (GCM) was coded as $-1$ and the prototype-based model (PM) was coded as 1. We report the coefficient $\hat{\beta}$ for the interaction term in the main paper. We found that ERMI ($\hat{\beta}=-0.01 \pm 0.004; z=-2.54,~p<0.01$) is better explained by the exemplar-based model with learning whereas choices from MI are explained equally well by exemplar-based and prototype-based learning ($\hat{\beta}=-0.002 \pm 0.005; z=-0.47,~p=0.63$) as shown in Figure \ref{fig:experiment2} 
     
    \paragraph{Prototype model cannot learn exceptions:} Given that the category structure is non-linearly separable (containing exceptions), a prototype model cannot explain the data fully even if provided the true category choices as it tends to miscategorize the exception stimulus from which category. The exemplar-based model (GCM) model, however, has no such issues and can fit the true choices perfectly. 
    
    \paragraph{MI and PFN models find it hard to learn exceptions:} A better fit of the prototype model to the MI and PFN in the latter half of the experiment (as shown in Figure \ref{fig:experiment2} and \ref{supp:experiment2}) suggests that, as observed in Shepard's task, they are not able to learn exceptions like the prototype model as mentioned before.

    \subsection{\textsc{Claude-v2} as a cognitive model of human category learning}

     \paragraph{Simulations:} To run simulations of the \citet{smith1998prototypes} study on \textsc{Claude-v2}, we queried it using the prompt shown in Appendix \ref{prompt:devraj2022}. The nonsense stimuli used in the experiment are provided in text. The order of presentation of the stimulus was also randomized, and the number of stimulus presentations in a block was matched to the original study.  We ran the model for a total of 120 simulation runs. 

    \begin{tcolorbox}[sharp corners, colback=mBlue!5!white,colframe=mBlue!75!black, width=1.\textwidth, left=4pt,right=4pt, top=4pt, bottom=4pt, title=\textbf{Prompt for  Devraj et al. 2022 study} \label{prompt:devraj2022}]
      In this experiment, you will be shown examples of nonsense word stimuli.  Look carefully at each word and decide if it belongs to group \{U\} or group \{M\}. Respond with \{U\} if you think it is a group \{U\} word and \{M\} if you think it is a group \{M\} word. You will recieve feedback about the correct group after each of your response.  At first, the task will seem quite difficult, but with time and practice, you should be able to answer correctly. \\
      
      - In trial 1, you picked group \{M\} for wafuzi and group \{U\} was correct.\\
      - In trial 2, you picked group \{M\} for gyfuzi and group \{U\} was correct.\\
      
      Human: What group would the word gyfuzi belong to? (Give the answer in the form ``Group $\langle$your answer$\rangle$").

Assistant: Group
    \end{tcolorbox}

    \begin{figure*}[!h]
    \centering
    \includegraphics[width=\textwidth]{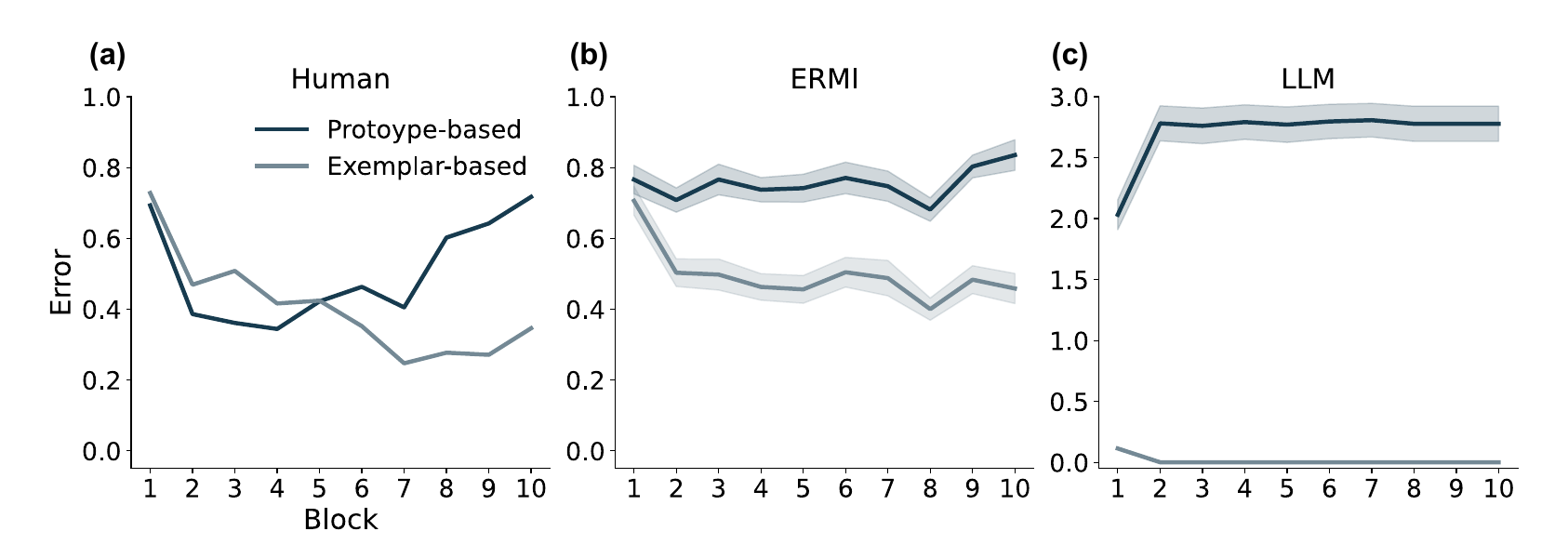}
    \caption{\textbf{\textsc{Claude-v2} performs purely exemplar-based category learning}: \textbf{(a-c)} The average error of exemplar- and prototype-based models fitted to (a) human choices, (b) simulated choices from ERMI, and (c) simulated choices from \textsc{Claude-v2} for each block of 56 trials. Human data in (a) was reproduced from \citet{smith1998prototypes}. ERMI was simulated using inverse temperature values fitted to participants' choices in \citet{devraj2021dynamics}. The mean of the fitted inverse temperature and its standard error for ERMI was $0.09 \pm 0.01$. \textsc{Claude-v2} was queried using the prompt shown in Appendix \ref{prompt:devraj2022} with the temperature term set to 0 for 120 runs. The shaded region shows the standard error of the mean.}
    \label{fig:llmsimulationsdevraj2022}
    \vspace{-0.20cm}
    \end{figure*}

\section{ERMI shows human-like generalization} \label{app:experiment3}

    \subsection{Experiment details for \citet{Johansen2002-xe}}
    \citet{Johansen2002-xe} conducted their study with 198 participants (out of which 68 were excluded for further analysis) using four-dimensional stimuli with each dimension taking one of two possible values. The stimuli were \say{computer-generated drawings of rockets that varied along four binary-valued dimensions: The shape of the wing (triangular or rectangular), tail (jagged or boxed), nose (staircase or half-circle), and porthole (circular or star)} \cite{Johansen2002-xe}. 
    The category structure used in the study was similar to the ones used in classical studies such as \citet{Medin1978-xf, Nosofsky1994-gu} and is ill-defined in that no single feature along a dimension can be used to perfectly classify stimuli. Rather, the categories have a family resemblance structure in that category A stimuli tend to have a value of  0 along each dimension, and category B stimuli tend to have a value of 1 along each dimension. In this case, category 1 contained five stimuli as follows: [0001, 0101, 0100, 0010, 1000]. The remaining four stimuli belonged to category 2 [0011, 1001, 1110, 1111]. The stimulus presentation order was randomized within each block.  Participants had unlimited time to make their choice on each trial. After making their choice, they were told whether or not it was a correct choice. Participants completed a total of 288 training trials, or 32 blocks of 9 trials each, in which they saw each stimulus once. However, in addition to the training block, participants had to perform a transfer block after 2, 4, 8, 16, 24, and 32 blocks of training. In a transfer block, all 16 possible stimuli are shown without any corrective feedback. 
    
    \subsection{Simulations} 
\begin{figure*}[!h]
     \centering
     \includegraphics[width=0.95\textwidth]{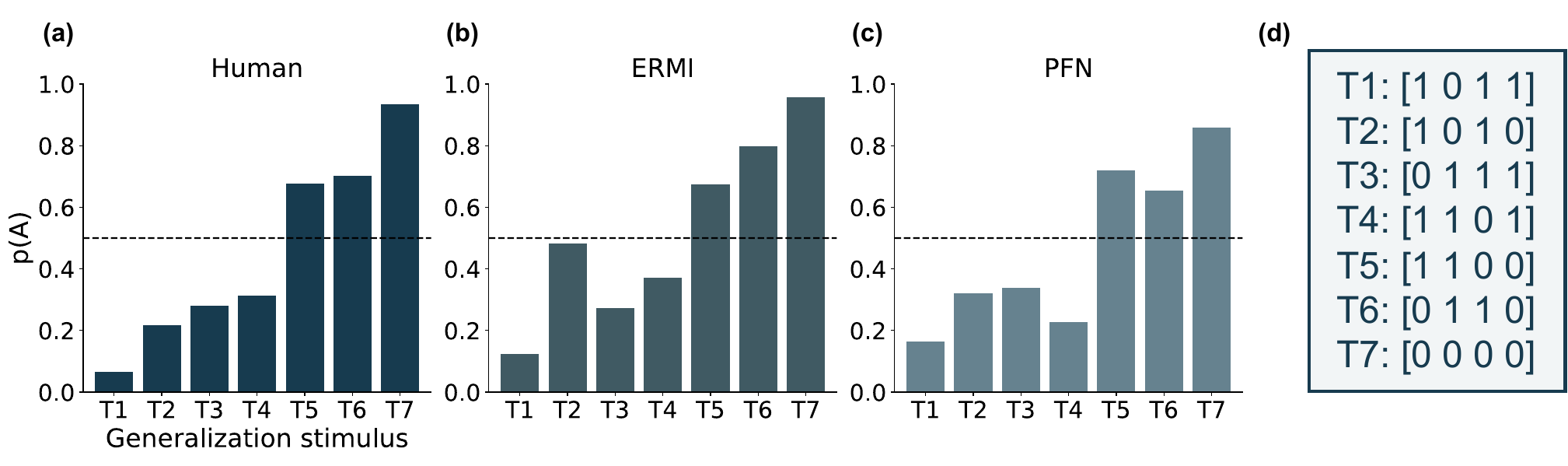}
      \caption{\textbf{Supplementary material accompanying Figure \ref{fig:experiment3}:} \textbf{(a-c)} Average categorization probabilities of transfer stimuli T1-T7 for (a) humans (b) ERMI (c) PFN. \textbf{(d)} The encoding scheme used for the seven transfer stimuli is provided for reference. Human data in (a) was reproduced from \citet{Johansen2002-xe}. ERMI and MI were simulated on the same experiment for 77 runs, with inverse temperature settings that resulted in the lowest mean-squared error compared to humans, which was $\beta=$ 0.9 for ERMI, and $\beta=$0.1 for PFN.}
     \label{supp:experiment3}
     \end{figure*}

We simulated ERMI, MI, and PFN on the \citet{Johansen2002-xe} study for different betas values, from zero to one in steps of 0.1, for a total of 544 runs. The models interacted with each of the nine training stimuli 32 times with the ordering of the stimuli shuffled between runs. Predictions for the transfer stimuli were derived by concatenating them -- one at a time -- at the end of 32 training blocks in every run. By doing so, we were able to derive the model's prediction for each unseen stimulus around $77$ times.  In Figure \ref{fig:experiment3} and \ref{supp:experiment3}, we reported average choice probabilities for the models using the $\beta$-term that minimized the pair-wise Euclidean distance between the human and model's choice probabilities. 

\begin{table}[!t]
    \caption{Detailed performance metrics of different models on OpenML-CC18 benchmarking suite. \\ 
    Abbreviations: Software defect prediction (SDP) and Service-Center (SC)}
    \label{tab:performance_metrics}
    \vskip 0.15in
    \begin{center}
        \begin{small}
            \begin{sc}
                \begin{tabular}{lccccc}
                \toprule
                \textbf{Data set} & \textbf{Log. Reg.} & \textbf{SVM} & \textbf{XGBoost} & \textbf{TabPFN} & \textbf{ERMI} \\
                \midrule
                kr-vs-kp  & 0.8257 & 0.8514 & 0.7986 & \textbf{0.8664} & 0.8450 \\ 
                credit-g  & \textbf{0.6421} & 0.6357 & 0.6350 & 0.6036 & 0.6150 \\ 
                diabetes  & 0.6771 & \textbf{0.7079} & 0.6786 & 0.6886 & 0.6950 \\ 
                spambase  & 0.5407 & 0.7664 & 0.7536 & \textbf{0.7993} & 0.7757 \\ 
                tic-tac-toe  & 0.5536 & 0.5950 & \textbf{0.6071} & 0.5914 & \textbf{0.6071} \\ 
                electricity  & 0.5543 & 0.6007 & \textbf{0.7036} & 0.6871 & 0.6436 \\ 
                pc4-sdp  & 0.7136 & 0.7521 & \textbf{0.7886} & 0.7707 & 0.7714 \\ 
                pc3-sdp & 0.6514 & 0.7264 & 0\textbf{.7357} & 0.7279 & 0.7107 \\ 
                kc2-sdp & 0.5893 & 0.7314 & \textbf{0.7257} & \textbf{0.7257} &\textbf{ 0.7257} \\ 
                kc1-sdp & 0.6271 & 0.6707 & \textbf{0.6743} & 0.6679 & 0.6521 \\ 
                pc1-sdp & 0.5336 & 0.5964 & 0.6514 & 0.6064 & 0.6493 \\ 
                wdbc  & 0.9121 & 0.9207 & 0.9014 & \textbf{0.9221 }& 0.9093 \\ 
                phoneme  & 0.5793 & \textbf{0.7314 }& 0.6979 & 0.6921 & 0.7200 \\ 
                qsar-biodeg  & 0.5779 & 0.7014 & 0.6850 & 0.6921 & \textbf{0.7064} \\ 
                ilpd  & 0.5493 &\textbf{ 0.6386} & 0.6229 & 0.6121 & 0.6286 \\ 
                ozone-level-8hr  & 0.6614 & 0.6907 & 0.6707 & 0.6471 & \textbf{0.6950 }\\ 
                banknote-authentication  & 0.7721 & 0.9229 & 0.8457 &\textbf{ 0.9657 }& 0.9379 \\ 
                blood-transfution-sc & 0.4714 & 0.5493 & 0.5879 & 0.5671 & \textbf{0.6186} \\ 
                phishing websites  & 0.7929 & 0.8071 & \textbf{0.8157 }& \textbf{0.8157} & 0.8129 \\ 
                bank-marketing  & 0.5829 & 0.5614 & \textbf{0.7386} & 0.7350 & 0.7171 \\ 
                wilt  & 0.5171 & 0.5736 & 0.6393 & 0.6371 & \textbf{0.6507} \\ 
                numerai28.6  & 0.4857 & 0.4779 & \textbf{0.5029} & 0.4779 & 0.4986 \\ 
                churn  & 0.6321 & 0.7271 & 0.6800 & 0.7186 & \textbf{0.7329 }\\     \midrule
                \textbf{Mean Acc.} & $62.80 \pm 0.66$ & $69.29\pm 0.62$ & 70.17$\% \pm 0.52$ & $70.51\% \pm 0.63$ & \textbf{70.95$\% \pm$ 0.54}\\    \midrule
                \textbf{Mean rank} & $4.52 \pm 0.21$  &  $2.76 \pm 0.26$ & $2.61 \pm 0.30$ &   $2.85 \pm 0.27$ &  \textbf{2.26 $\pm$ 0.22}\\     
                \bottomrule
                \end{tabular}
            \end{sc}
        \end{small}
        \end{center}
    \vskip -0.1in 
\end{table}
\section{Benchmarking on OpenML-CC18} \label{app:benchmark}
Table \ref{tab:performance_metrics} contains the full set of results for all tasks and models.
    
\section{Software and Data}

We have made the data and code available under the following link: \url{https://github.com/akjagadish/ermi}



\end{document}